# Experimental Analysis of Type II Singularities and Assembly Change Points in a 3UPS+RPU Parallel Robot

Pulloquinga José L.[a*], Mata Vicente[b], Valera Ángel[a], Zamora-Ortiz Pau[b], Díaz-Rodríguez Miguel[c], Zambrano Iván[d]

[a] *Universitat Politècnica de València, Departamento de Ingeniería de Sistemas y Automatica, Instituto de Automática e Informática Industrial, Valencia-Camino de Vera, s/n, Spain. Email: jopulza@doctor.upv.es, giuprog@isa.upv.es*
[b] *Universitat Politècnica de València, Departamento de Ingeniería Mecánica y de Materiales, Centro de Investigación en Ingeniería Mecánica, Valencia-Camino de Vera, s/n, Spain. Email: vmata@mcm.upv.es, pazaor@doctor.upv.es*
[c] *Universidad de los Andes, Núcleo la Hechicéra, Departamento de Tecnología y Diseño, Facultad de Ingeniería, 5101, Merida, Venezuela. Email: dmiguel@ula.ve*
[d] *Escuela Politécnica Nacional, Departamento de Ingeniería Mecánica, Facultad de Ingeniería Mecánica, Quito-Ladrón de Guevara E11-253, Ecuador. Email: ivan.zambrano@epn.edu.ec*

**Abstract**

Parallel robots (PRs) have singular configurations where the robot gains at least one degree-of-freedom and loses control. Theoretically, such singularity occurs when the Forward Jacobian-Matrix determinant becomes zero (Type II). However, actual PRs could lose control owing to Type II singularities for determinant values near zero, but not zero, because manufacturing tolerances introduce errors that are complex to model due to their low repeatability. Thus, using an actual 3UPS+RPU PR, this paper presents three contributions: i) a proximity detection index for Type II singularities based on the angle between two Output Twist Screws. The index can identify which kinematic chains contribute to the singularity. ii) an experimental benchmark to study Type II singularities. iii) PR configurations where the proposed index is zero and the Forward Jacobian determinant is not. In this last configuration, the findings show that the actual robot is unable to handle external actions applied to the PR.

## 1. Introduction

In contrast to a serial robot, a parallel robot (PR) drives its end-effector (mobile platform) using at least two closed kinematic chains that help increase the PR stiffness and load capacity. Moreover, PRs have other advantages over their serial counterparts, such as lower weight, higher working speed with high precision, and lower power consumption [1,2]. These advantages, mainly due to the closed kinematic chain architecture, are key aspects that have increased the interest in studying their use in the academic, industrial, and robotics service fields over the last three decades. However, the PR architecture reduces not only the size of the robot workspace but also its kinematic performance, owing to the possible presence of singularities within the workspace.

Initially, Gosselin and Angeles [3] studied the singularities of a PR using Jacobian matrices obtained from constraint equations, and classified them into i) inverse kinematic or Type I singularity, where the robot loses at least one degree of freedom (DOF), and ii) Forward Kinematic or Type II singularity, where the PR gains at least one DOF. In particular, Type II singularities could be critical because the mobile platform at the singularity is unable to bear the external forces despite having all the actuators locked (losing control of the PR motion). Likewise, Park and Kim [4], based on a similar Jacobian matrix analysis, classified PR singularities as actuator (analogous to Type I) and end-effector (equivalent to Type II) singularities. Gregorio and Parenti-Castelli, in [5], subclassified Type II singularities using a 3-UPU PR according to the type of DOF gained (rotational or translational). Moreover, Slavutin et al. [6] proposed a graphical analysis for a spatial PR using the three-dimensional Kennedy theorem.

The singularities of a PR, due to the instantaneous change in its DOF, are a problem to avoid. The conventional way to detect that a PR undergoes a singular configuration is by evaluating the determinant of the Jacobian matrix. When the determinant is zero, the PR is in a singular configuration. The Inverse Jacobian matrix is analyzed for a Type I singularity ($J_I$) and the Forward Jacobian matrix for a Type II singularity ($J_D$). However, the determinant of the Jacobian matrix ($||J||$) does not have physical meaning [7]. Merlet, in [8], proposed alternative ways of detecting singularities based on the Jacobian matrix concept: manipulability index, condition number, and global condition index. Gallardo et al. [9] calculate a Jacobian Matrix based on Screw Theory and the Principle of Virtual Work to detect singularities on a Schönflies PR. Some of these indices have an unclear physical meaning in some cases and have not been applied to actual robots.

---

\* Corresponding author. E-mail: jopulza@doctor.upv.es

An alternative way to deal with singularities is to design the PRs focusing on optimizing the workspace to avoid the singular configuration. Davidson et al. [10] introduced an optimization procedure based on force/motion transmission efficiency. Yuan et al. [11] used Screw Theory to define a transmission index between the Transmission Wrench Screw (TWS) of the actuators and their Output Twist Screw (OTS) produced in the mobile platform. Chen and Angeles [12] established a general transmission index (GTI) considering the TWS as a general transmission wrench screw (GTWS). Takeda and Funabashi [13] defined a transmission index that was able to establish how each actuator contributes to the motion of the mobile platform; for this purpose, Ref. [13] considered just one actuator as active while the others were locked. Subsequently, the transmission index was normalized by Wang et al. [14], dividing it by the maximum virtual power of each actuator; the normalized index was named Output Transmission Index (OTI). Araujo-Gómez et al. [15] showed that even after optimizing the workspace of a 4 DOF PR, a small percentage of singularities still remain inside the workspace of the robot. Therefore, in a lower-mobility PR with a singular configuration inside its workspace, it becomes necessary to determine when the PR is near to a singular configuration. Hesselbach et al. [16] predicted the proximity of a singularity through online optimization algorithms (based on integrated sensors) to minimize robot configurations with zero transmitted power.

Hunt and Primrose [17], on the other hand, initially indicated that within the workspace free of Type II singularities, a solution for the Forward Kinematic problem or assembly mode [18] is unique. Thus, to move from one assembly mode ($\|J_D\| > 0$) to another ($\|J_D\| < 0$) it is necessary to undergo a Type II singularity ($\|J_D\| = 0$). Later, Innocenti and Parenti-Castelli [19] showed that for PRs, there could be several solutions to the Forward Kinematic problem. This introduced the possibility of switching from one assembly mode to another without passing through a Type II singularity ($\|J_D\|$ does not pass through zero). A very effective way to perform a non-singular assembly change is to go through a special Type II singularity called cuspidal point [20]. In [21], based on series expansion, McAree and Daniel defined a cuspidal point as a second-order degeneration or a special case of Type II singularity where three assembly modes are certain to meet. References [22–24] presented algorithms for trajectory planning avoiding singular points.

Generally, the singularities of a PR are analyzed without modeling manufacturing errors because of their random behavior. Dali et al., in [25], showed that the actual position of the mobile platform of a PR has errors caused by manufacturing errors in links and joints. Chen et al. [26] model the position error due to joint clearances and input uncertainties in different joints of a planar PR. The study verified that, for each position of the planar PR, the effect of joint clearances is not deterministic. Binaud et al. [27], emphasized the difficulty of correcting the positioning errors due to joint clearances because of their low repeatability. Huang et al. [28] estimate the position error due to manufacturing errors using a probabilistic model. Moreover, Ohno and Takeda in [29] used embedded sensors to estimate the position error due to clearances.

A literature search revealed that few studies have analyzed the closeness to a singularity in spatial PRs from an experimental perspective. Therefore, the contributions of this study are presented below:

i) Proposes to use the angle ($\Omega$) between two instantaneous screw axes from the Output Twist Screws (OTSs) as an index capable of detecting the proximity to Type II singularities. OTSs were used in [13] as part of the design optimization for PRs. The main advantages of the index $\Omega$ are physical meaning, sensitivity near a singularity and identification of the robot kinematic chains producing the singular configuration.

ii) Presents an experimental benchmark between the index $\Omega$ and $\|J_D\|$ to analyze Type II singularities in an actual 3U$\underline{P}$S+R$\underline{P}$U PR. The experiments show that the PR approaches a singular configuration for $\|J_D\|$ different from zero, i.e. $\|J_D\|$ identifies a singularity in the vicinity of zero. The experimental benchmark starts measuring the position of the mobile platform using a high-precision photogrammetry system. Subsequently, based on the measured positions, the values of $\|J_D\|$ and $\Omega$ are calculated and used to set the non-zero experimental limits to ensure that the actual PR avoids a Type II singularity. The set of trajectories employed to calculate the experimental limits relies on typical knee rehabilitation and diagnosis tasks. Finally, to verify that the $\|J_D\|$ and $\Omega$ limits were correctly established, several trajectories where the $\|J_D\|$ is very close to the limit are executed. These verification trajectories also allow to show that the proposed index is more sensitive and effective than the $\|J_D\|$ in detecting the proximity of Type II singularities for an actual PR. The proposed benchmark is not intended to consider joint clearances model explicitly. However, this experimental protocol can obtain limits for $\|J_D\|$ and $\Omega$ that are effective in avoiding singular configuration even under manufacturing errors in links and joints. The proposed protocol could be applied when developing an actual PR to obtain a safety index for avoiding singular configurations.

iii) Introduces PR configurations where, in simulation, the angle $\Omega$ is zero, despite being non-singular or cuspidal points. In this special configuration, the actual analyzed PR loses motion control and allows it to undergo from one assembly mode to another. Therefore, this paper calls these points assembly changing points (AC points). Using simulated and experimental tests, the index $\Omega$ shows its capacity to detect AC points that cannot be detected using $\|J_D\|$.

Section 2 presents a briefly description of the 3UPS+RPU PR, singularities and cuspidal points in PRs, and the angle between the OTSs. Section 3 describes the procedure used to determine the $\|J_D\|$ and the indices $\Omega$. The section also introduces the expression defining the number of assembly modes for the 3UPS+RPU PR that are used to verify the assembly change performance by the actual PR in an AC point. Moreover, it describes the test trajectories used to set the experimental limits for the $\|J_D\|$ and the $\Omega$ that avoid Type II singularities. Moreover, Section 3 describes the verification test to ensure the previously set experimental limits and the trajectories with AC points where the actual PR cannot hold external actions. Section 4 presents the test-bed for the experimental benchmark to analyze Type II singularities. The obtained results are also shown and discussed in this Section. Finally, in Section 5 the main conclusions are presented.

## 2. Mathematical foundation

### 2.1. 3UPS+RPU parallel robot

In the 3UPS+RPU PR (see Fig. 1a) the fixed base is linked to the end-effector or mobile platform by means of four open kinematic chains or limbs, in which three of them have a UPS configuration and the fourth one has an RPU configuration. The letter R stands for revolute joint, P for prismatic joint (actuated in this case, indicated by the underlined format), U for universal joint, and S for spherical joint. The 3UPS+RPU robot was designed and built at Universitat Politècnica de València. The robot has four DOF: two translational and two rotational. The kinematic model of this PR and its optimization was developed in [15,30,31].

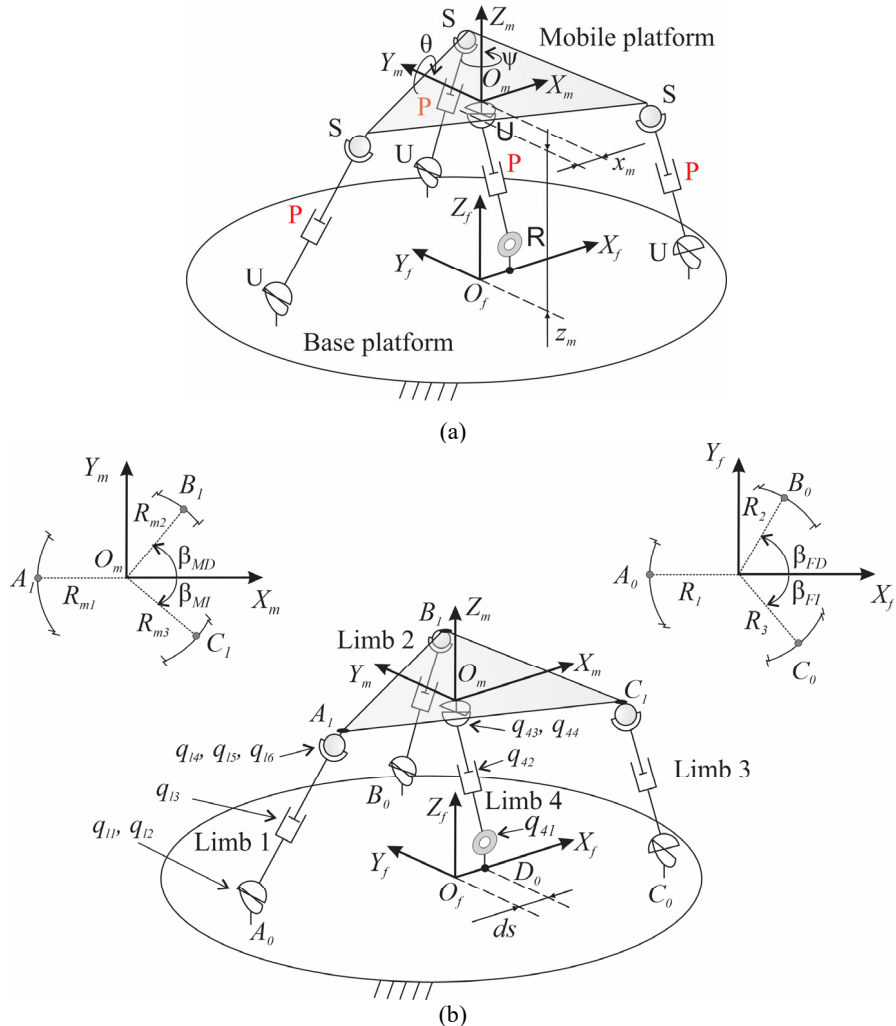

Fig. 1. 3UPS+RPU-type PR (a) Simplified representation (b) Mechanical configuration.

Two reference systems $\{O_f - X_F Y_F Z_F\}$ and $\{O_m - X_M Y_M Z_M\}$, see Fig. 1b, are attached to the fixed and mobile platforms of the PR, respectively. The location of reference system attached is at the center of the mobile platform as this location corresponds to the contact point with the patient's leg. In contrast, the reference system of the fixed platform is established at the center of the platform to reduce model burden and facilitate the measure of the mobile platform position.

The position of the mobile platform is given by $x_m, y_m, z_m$, with $y_m = 0$, because the R joint of the central limb (4) constrains the movements on the $Y_F$ axis. In addition, the U joint linking limb 4 with the mobile platform only allows two rotations, one around the $Y_M$ axis represented by $\theta$ and another around the $Z_M$ axis, defined by $\psi$.

To derive the kinematic model, the Denavit-Hartenberg convention is considered [32]. In Fig. 1b, the universal joint in limbs $l = 1 \ldots 3$ is modeled as two perpendicular revolute joints, $q_{l1}$ and $q_{l2}$ being the generalized coordinates. The displacement of the prismatic joint corresponds to $q_{l3}$ and the spherical joints are modelled as three mutually orthogonal revolute joints with $q_{l4}$, $q_{l5}$ and $q_{l6}$ as generalized coordinates. For the central limb (limb 4), $q_{41}$ is the generalized coordinate corresponding to the revolute joint, its prismatic displacement is modeled by $q_{42}$, and the generalized coordinates corresponding to the universal joint are $q_{43}$ and $q_{44}$. The location of the point $A_0$ on the fixed platform is defined by the radius $R_1$ measured from $O_f$ and parallel to $X_F$ axis. The $B_0$ is determined using radius the $R_2$ measured from $O_f$, and the angle $\beta_{FD}$ measured counterclockwise from $X_F$ axis. The location of $C_0$ is characterized by the radius $R_3$ and the angle $\beta_{FI}$ measured clockwise from $X_F$ axis. On the mobile platform, the link points to the external limbs ($A_1$, $B_1$, $C_1$) are defined similarly to $A_0$, $B_0$, $C_0$ by using $R_{m1}, R_{m2}, R_{m3}, \beta_{MD}$ and $\beta_{MI}$. Regarding limb 4, the location of $D_0$ on the fixed platform is defined by the distance $d_s$ measured from $O_f$. On the mobile platform, the limb 4 is connected directly to the origin $O_m$ coordinate system.

In Table 1, the parameters used for the 3UPS+RPU PR are listed, both for simulation an experimentation.

Table 1: Parameters for 3UPS+RPU PR.

| $R_1$ (m) | $R_2$ (m) | $R_3$ (m) | $\beta_{FD}$ (°) | $\beta_{FI}$ (°) | $d_s$ (m) |
|---|---|---|---|---|---|
| 0.4 | 0.4 | 0.4 | 90 | 45 | 0.15 |
| $R_{m1}$ (m) | $R_{m2}$ (m) | $R_{m3}$ (m) | $\beta_{MD}$ (°) | $\beta_{MI}$ (°) | |
| 0.3 | 0.3 | 0.3 | 50 | 90 | |

*2.2. Singularities*

In a closed kinematic chain robot, a set of constraint equations, $\vec{\Phi}$, defines the relationship between inputs or active generalized coordinates, $\vec{q}_{ind}$, and outputs or degrees of freedom (DOF) of the mobile platform ($\vec{X}$) [3].

$$\vec{\Phi}(\vec{X}, \vec{q}_{ind}) = \vec{0} \tag{1}$$

Taking time derivatives, the velocity equations for a PR are as follows:

$$J_D \dot{\vec{X}} + J_I \dot{\vec{q}}_{ind} = \vec{0} \tag{2}$$

where $J_D$ is the Forward Jacobian matrix and $J_I$ is the Inverse Jacobian matrix. For a non-redundant PR both matrices are $FxF$ square matrices, $F$ being the DOF of the mobile platform.

A Type I singularity occurs when the $J_I$ matrix is rank deficient $\|J_I\| = 0$. In this case, the mobile platform cannot move ($\dot{\vec{X}} = 0$) despite having a set of non-zero velocities in the actuators; in other words, at least one DOF is lost. In [2] this type of singularity is analyzed in detail.

A Type II singularity that corresponds to the Forward Kinematic problem is defined by isolating $\dot{\vec{X}}$ in (2), thereby:

$$\dot{\vec{X}} = -J_D^{-1} J_I \dot{\vec{q}}_{ind} \tag{3}$$

In this case, the mobile platform can perform some motion, $\dot{\vec{X}} \neq 0$, if an external action is applied to the mobile platform, despite any motion in the actuators, $\dot{\vec{q}}_{ind} = 0$. That is, the mobile platform gains at least one DOF ($\|J_D\| = 0$). Under these conditions, control over the robot is lost, which is potentially dangerous for the user or the robot itself. In [33] these types of singularities are analyzed using a 2PRU+1PRR robot.

A Type II singulary implies first-order degeneration where at least two assembly modes are certain to meet [21]. In planar PRs, a Type II singularity where three assembly modes match [21], the terms associated with $\partial^2 \vec{\Phi} / \partial \vec{X}^2$ must satisfy:

$$v_3^T \left[ u_{13} \frac{\partial^2 \vec{\Phi}_1}{\partial \vec{X}^2} + u_{23} \frac{\partial^2 \vec{\Phi}_2}{\partial \vec{X}^2} + u_{33} \frac{\partial^2 \vec{\Phi}_3}{\partial \vec{X}^2} \right] v_3 = 0 \tag{4}$$

where $u_{13}, u_{23}, u_{33}$ are elements of a row vector, and $v_3$ is a column vector, both vectors are nonzero and extracted from $adj\left(\frac{\partial \vec{\Phi}}{\partial \vec{X}}\right)$, where $adj(\ )$ is the adjoint operator.

For planar PRs, Type II singularities with second-order degeneration are called cuspidal points because their position projects a cusp on the joint plane (Fig. 2). For spatial PRs, due to the higher number of DOF and actuators, the representation of these points is not possible, but the name cuspidal has been retained. A cuspidal point is used as a pivot in a non-singular assembly mode change because it is a convergence point of the assembly modes, and the PR will not undergo a Type II singularity around it.

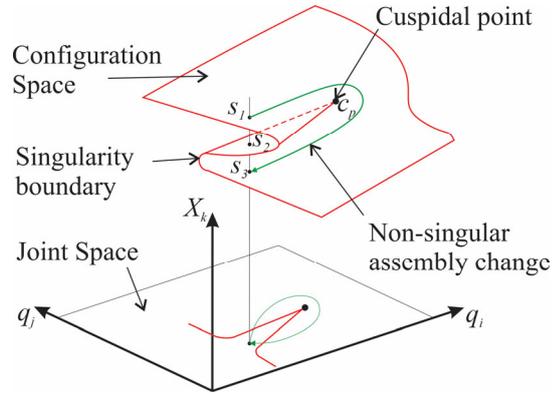

Fig. 2. Cuspidal points and non-singular assembly mode changing.

*2.3. Angle between Output Twist Screws*

In a PR, the motion of the mobile platform is produced by the combined action of several actuators, which renders it difficult to identify the individual contribution of each actuator.

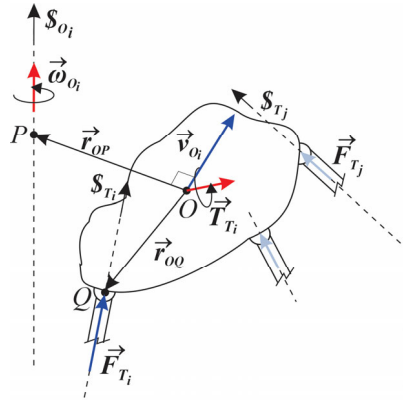

Fig. 3. Instantaneous power from one actuator.

Takeda and Funabashi, in [13], proposed to separate the motion of a point of the mobile platform in $F$ OTSs $\$_O$. For this purpose $F$ actuators are locked except one (see Fig. 3), i.e. the mobile platform's movement ($\$$) can be expressed as:

$$\$ = \rho_1 \hat{\$}_{O_1} + \rho_2 \hat{\$}_{O_2} + \cdots + \rho_F \hat{\$}_{O_F} \tag{5}$$

where $\rho_i$ is the amplitude for each OTS and $\hat{\$}_{O_i}$ is a normalized OTS.

To determine the $F$ normalized OTSs ($\hat{\$}$), the instantaneous power produced by the locked actuators must be analyzed, using the expression:

$$\hat{\$}_{O_i} \circ \hat{\$}_{T_j} = 0 \quad (i, j = 1, 2, \ldots, F, i \neq j) \tag{6}$$

where $\circ$ stands for the reciprocal product and $\hat{\$}_{T_j}$ is the unitary screw of each TWS of the actuators that are considered locked.

Wang et al., in [14], proved that for a non-singular configuration the $F$ $\hat{\$}_O$ are linearly independent. Then for a singular configuration at least two $\hat{\$}_O$ are linearly dependent. Thus,

$$\hat{\$}_{O_i} = \hat{\$}_{O_j} \tag{7}$$



with:

$$\hat{\$}_O = (\vec{\mu}_{\omega_O}; \vec{\mu}_{v_O}^*) = (\vec{\mu}_{\omega_O}; h\vec{\mu}_{\omega_O} + \vec{r} \times \vec{\mu}_{\omega_O}) \tag{8}$$

where $\vec{\mu}_{\omega_O}$ is the instantaneous screw axis, $\vec{r}$ is the minimal distance between the moving element and $\vec{\mu}_{\omega_O}$, and $h$ is the screw's pitch. For a more detailed analysis of Screw Theory the reader may refer to [10].

In other words, in a Type II singularity both angular ($\vec{\mu}_{\omega_O}$) and linear components ($\vec{\mu}_{v_O}^*$) of two $\hat{\$}_O$ are equal. As $\vec{\mu}_{\omega_O}$ is a unit vector, the equality between two different angular components is defined by the parallelism between them. Based on this property, this paper proposes to analyze Type II singularities considering pairs of $\hat{\$}_O$. The proximity to a Type II singularity is measured by the angle of two $\vec{\mu}_{\omega_O}$, named $\Omega$. Moreover, the proximity is verified by the equality of $\vec{\mu}_{v_O}^*$. Considering $F$ $\hat{\$}_O$ grouped in pairs, there are $\binom{F}{2}$ angles $\Omega$, that is,

$$\Omega_{i,j} = acos\left(\vec{\mu}_{\omega_{O_i}} \cdot \vec{\mu}_{\omega_{O_j}}\right) \quad (i,j = 1,2,\dots,F, i \neq j) \tag{9}$$

where $i$ and $j$ identify the selected $\hat{\$}_O$.

From a theoretical perspective, a PR undergoes a Type II singularity, if and only if $\|J_D\| = 0$, $\Omega_{i,j} = 0$ and $\vec{\mu}_{v_{O_i}}^* = \vec{\mu}_{v_{O_j}}^*$. This study considers both components of $\hat{\$}_O$, however the analysis is limited to study on the angular component between two OTSs.

As an index $\Omega_{i,j}$ is an angular measure, the proximity to a Type II singularity is measured in angular units, unlike the $\|J_D\|$. This physical meaning of a $\Omega_{i,j}$ allows to identify the chains that cause a Type II singularity. Note that when $\Omega_{i,j} = 0$, $\hat{\$}_{O_i}$ and $\hat{\$}_{O_j}$ degenerate in expression (6), implying that the pair of chains $i, j$ are responsible for the singular configuration. Thus, further action can be taken based on this index, for instance, to develop a strategy for reconfiguring the pair of chains responsible for the singular configuration.

A $\hat{\$}_{O_i}$ was defined by Takeda and Funabashi in [13] to obtain the Output Transmission Indices (OTIs) used for the optimal design of PRs. These OTIs are applied in Ref. [34] to classify singularities on a Schönflies PR; however, the $\hat{\$}_O$ is not analyzed considering each component. Then, a $\hat{\$}_O$ has not been used directly in the detection of Type II singularities, which means that an angle $\Omega_{i,j}$ has not been used for analyzing Type II singularities. Therefore, the first contribution of this study is the angle $\Omega_{i,j}$ as a proximity detection index for Type II singularities. In the next section the indices $\Omega_{i,j}$ for the PR under study are developed.

## 3. Simulations

This Section first presents a procedure to obtain the $J_D$ for a 3U$\underline{P}$S+R$\underline{P}$U PR using the constraint equations. Subsequently, each $\Omega_{i,j}$ index is defined by analyzing the OTS produced by the four TWSs corresponding to the actuators of the robot under study. Later, the equations that define the possible assembly modes for the 3U$\underline{P}$S+R$\underline{P}$U PR are developed. Then, three set of trajectories that are used in simulations to fulfill the three contributions of this study are presented, which are:

- Test Trajectories: Knee rehabilitation trajectories in the PR configuration space, some of which go through Type II singular configurations. The trajectories verify that in a Type II singular configuration the $\|J_D\|$ and the $\Omega_{i,j}$ index are zero and $\vec{\mu}_{v_{O_i}}^* = \vec{\mu}_{v_{O_j}}^*$, i.e. the proposed $\Omega_{i,j}$ index is able to detect the proximity to a Type II singularity. In this study, we evaluate all $\Omega_{i,j}$; nevertheless, only the angle closest to zero, which is the responsible for the singularity, is presented.
- Verification Trajectories: Trajectories that theoretically do not contain any Type II singular configuration. The movements performed have no knee rehabilitation purposes as opposed to those performed in the test trajectories. The trajectories start from a configuration with $\|J_D\| \gg 0$, and during the trajectories the $\|J_D\|$ decreases up to the final position. Based on these trajectories, the analysis of the average rate of change of both the $\|J_D\|$ and the $\Omega_{i,j}$ index shows that the index proposed decreases faster than the $\|J_D\|$ when the 3U$\underline{P}$S+R$\underline{P}$U PR gets closer to a Type II singularity configuration. In other words, a $\Omega_{i,j}$ index has more sensitivity in the proximity of a Type II singularity.
- Trajectories with assembly change points: There are trajectories whose end configuration has $\Omega_{i,j} = 0$ and $\|J_D\| \neq 0$. The particular points mentioned cannot be singular or cuspidal, because $\|J_D\| \neq 0$ and $\vec{\mu}_{v_{O_i}}^* \neq \vec{\mu}_{v_{O_j}}^*$.

    Later on, Section 4.4 shows experimentally that at these points the control of the mobile platform is lost, and it experimentally allows a singular or non-singular assembly change (passing from one direct kinematic solution to another). PR positions with these characteristics are referred to herein as assembly change points (AC points). The simulation of the trajectories with assembly change points shows that the $\Omega_{i,j}$ index is able



to detect AC points. The possibility of AC points being singular or cuspidal is eliminated using the $\|J_D\|$ and the linear term $\vec{u}_{v_O}^*$ of the pair $i, j$ unitary OTS.

*3.1. Forward Jacobian matrix*

To establish the equations (1), we consider the relationship of the closed chains between the position vector from the origin $O_f$ of the fixed reference system to the origin $O_m$ of the moving reference system and the open chains formed using the different limbs of the robot. The open chains take $O_f$ and $O_m$ as the starting and end point, respectively.

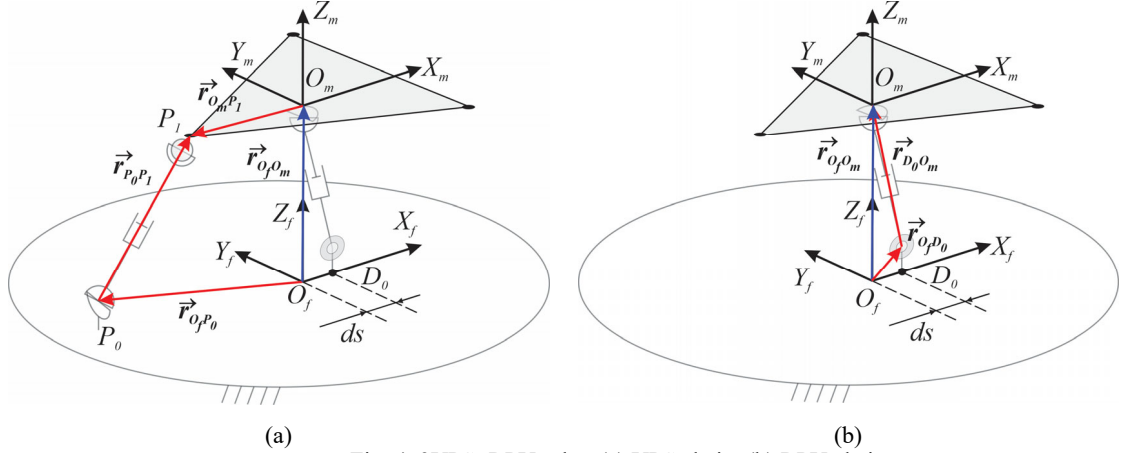

Fig. 4. 3U<u>P</u>S+R<u>P</u>U robot (a) UPS chain, (b) RPU chain.

For a UPS chain (Fig. 4a), we have:

$$|\vec{r}_{P_0 P_1}| = |\vec{r}_{O_f O_m} + \vec{r}_{O_m P_1} - \vec{r}_{O_f P_0}| \tag{10}$$

with $\mathcal{P} = A, B, C$ for limbs 1, 2 and 3, respectively.
For an RPU chain (Fig. 4b), we have:

$$|\vec{r}_{D_0 O_m}| = |\vec{r}_{O_f O_m} - \vec{r}_{O_f D_0}| \tag{11}$$

where $\vec{r}_{O_f O_m}$ is the position of the center of the mobile platform with regard to the fixed reference system.

$$\vec{r}_{O_f O_m} = [x_m \quad 0 \quad z_m]^T \tag{12}$$

Vectors $\vec{r}_{O_m P_1}$ are determined by the rotation matrix ${}^f R_m$ and the vectors from the origin $O_m$ to each vertex on the mobile platform with respect to the moving reference system.

$$\vec{r}_{O_m P_1} = {}^f R_m \, {}^m\vec{r}_{O_m P_1} \tag{13}$$

where ${}^f R_m$ is the rotation matrix between the moving and fixed reference system. Considering the Euler Y-Z' angle convention with respect to moving axes, then:

$${}^f R_m = \begin{bmatrix} \cos\theta\cos\psi & -\cos\theta\sin\psi & \sin\theta \\ \sin\psi & \cos\psi & 0 \\ -\sin\theta\cos\psi & \sin\theta\sin\psi & \cos\theta \end{bmatrix} \tag{14}$$

Developing (10) and (11), the set of constraint equations $\vec{\Phi}(\vec{X}, \vec{q}_{ind})$ is defined as:

$$\begin{aligned} & q_{13}^2 + a_1 C_\theta C_\psi + 2R_{m1} C_\theta C_\psi x_m - 2R_{m1} S_\theta C_\psi z_m - 2R_1 x_m - x_m^2 - z_m^2 + a_2 = 0 \\ & q_{23}^2 - b_1 C_\theta S_\psi + b_2 C_\theta C_\psi + b_3 C_\theta S_\psi x_m + b_4 S_\psi - b_3 S_\theta S_\psi z_m - b_5 C_\theta C_\psi x_m + b_6 C_\psi + b_5 S_\theta C_\psi z_m \\ & \quad + b_7 x_m - x_m^2 - z_m^2 + b_8 = 0 \end{aligned} \tag{15}$$



$$q_{33}^2 + c_1 C_\theta S_\psi + c_2 C_\theta C_\psi - c_3 C_\theta S_\psi x_m - c_4 S_\psi - c_5 C_\theta C_\psi x_m + c_6 C_\psi + c_3 S_\theta S_\psi z_m + c_5 S_\theta C_\psi z_m + c_7 x_m$$
$$- x_m^2 - z_m^2 + c_8 = 0$$

$$q_{42}^2 - ds^2 + 2ds x_m - x_m^2 - z_m^2 = 0$$

where:

| | | | |
|---|---|---|---|
| $C_\theta = \cos\theta$ | $a_1 = 2R_1 R_{m1}$ | $b_1 = 2R_2 R_{m2} C_{FD} S_{MD}$ | $c_1 = 2R_3 R_{m3} C_{FI} S_{MI}$ |
| $S_\theta = \sin\theta$ | $a_2 = -R_1^2 - R_{m1}^2$ | $b_2 = 2R_2 R_{m2} C_{FD} C_{MD}$ | $c_2 = 2R_3 R_{m3} C_{FI} C_{MI}$ |
| $C_\psi = \cos\psi$ | | $b_3 = 2R_{m2} S_{MD}$ | $c_3 = 2R_{m3} S_{MI}$ |
| $S_\psi = \sin\psi$ | | $b_4 = 2R_2 R_{m2} S_{FD} C_{MD}$ | $c_4 = 2R_3 R_{m3} S_{FI} C_{MI}$ |
| $C_{FD} = \cos(\beta_{FD})$ | | $b_5 = 2R_{m2} C_{MD}$ | $c_5 = 2R_{m3} C_{MI}$ |
| $S_{FD} = \sin(\beta_{FD})$ | | $b_6 = 2R_2 R_{m2} S_{FD} S_{MD}$ | $c_6 = 2R_3 R_{m3} S_{FI} S_{MI}$ |
| $C_{FI} = \cos(\beta_{FI})$ | | $b_7 = 2R_2 C_{FD}$ | $c_7 = 2R_3 C_{FI}$ |
| $S_{FI} = \sin(\beta_{FI})$ | | $b_8 = -R_2^2 - R_{m2}^2$ | $c_8 = -R_3^2 - R_{m3}^2$ |
| $C_{MD} = \cos(\beta_{MD})$ | | | |
| $S_{MD} = \sin(\beta_{MD})$ | | | |
| $C_{MI} = \cos(\beta_{MI})$ | | | |
| $S_{MI} = \sin(\beta_{MI})$ | | | |

If $\vec{X} = [x_m \quad z_m \quad \theta \quad \psi]^T$ and $\vec{q}_{ind} = [q_{13} \quad q_{23} \quad q_{33} \quad q_{42}]^T$, the $J_D$ is defined as a 4x4 matrix:

$$J_D = \begin{bmatrix} \frac{\partial \vec{\Phi}}{\partial x_m} & \frac{\partial \vec{\Phi}}{\partial z_m} & \frac{\partial \vec{\Phi}}{\partial \theta} & \frac{\partial \vec{\Phi}}{\partial \psi} \end{bmatrix} \tag{16}$$

### 3.2. Angle between two Output Twist Screws

To define every $\Omega_{i,j}$ index, it is necessary to establish four $\hat{\$}_O$ ($F = 4$) on a point in the mobile platform (see Fig. 5a). For the sake of simplicity, the analysis considers the origin of the mobile platform $O_m$. The $\hat{\$}_T$ corresponding to the four actuators are:

$$\hat{\$}_{T1} = \begin{bmatrix} \vec{z}_{12} \\ \vec{r}_{O_m A_1} \times \vec{z}_{12} \end{bmatrix}, \hat{\$}_{T2} = \begin{bmatrix} \vec{z}_{22} \\ \vec{r}_{O_m B_1} \times \vec{z}_{22} \end{bmatrix}, \hat{\$}_{T3} = \begin{bmatrix} \vec{z}_{32} \\ \vec{r}_{O_m C_1} \times \vec{z}_{32} \end{bmatrix}, \hat{\$}_{T4} = \begin{bmatrix} \vec{z}_{41} \\ \vec{0} \end{bmatrix} \tag{17}$$

where $\vec{z}$ is the unit vector in the direction of the force applied by the actuators and $\vec{r}$ the vector measured from $O_m$ to each vertex connecting the mobile platform with the limbs of the PR (see Fig. 5b).

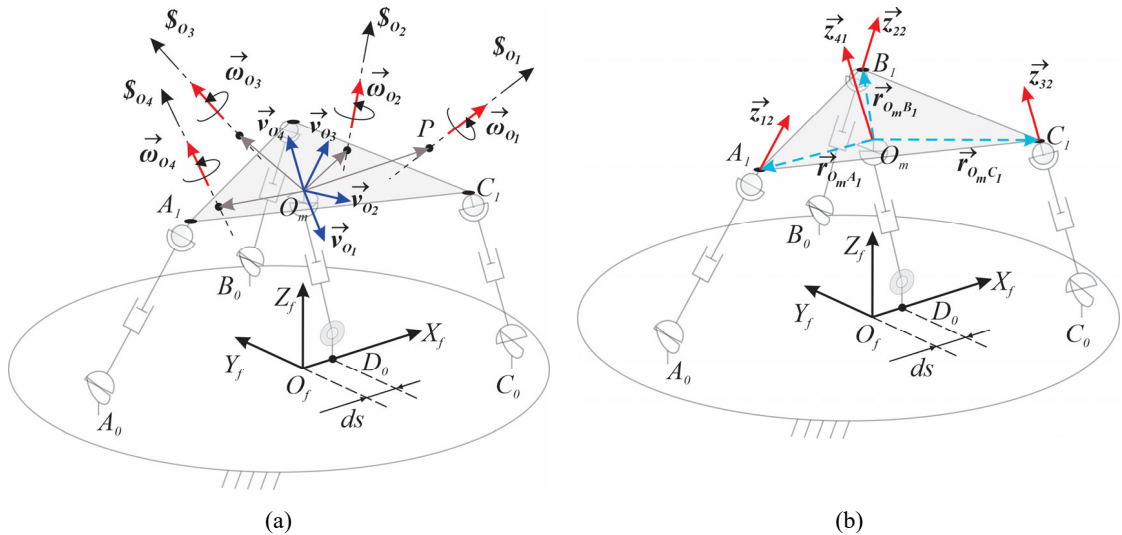

(a)        (b)

Fig. 5. Screws of the parallel robot (a) OTS, (b) TWS.

By replacing (17) in (6), a system of three equations to calculate four $\hat{\$}_O$ can be defined. There are several methods to solve a reciprocal screw of order $F$, as detailed explained in [35]. In this case, the screw system of $F$ $\hat{\$}_O$ is completed



based on the geometrical constrains of the 3UPS+RPU PR. Thus, for each $\hat{\$}_O$ there are five components ($\omega_{O_x}$, $\omega_{O_y}$, $\omega_{O_z}$, $v_{O_x}$ and $v_{O_z}$). Thus, two more equations are needed to complete the system. The component $v_{O_y} = 0$ because of the physical constraints of the robot in $O_m$, that is the point selected to analyze every $\hat{\$}_O$. The fourth equation is given by the norm of the instantaneous screw axis $\vec{\mu}_{\omega_O}$, as follows:

$$\omega_{O_x}^2 + \omega_{O_y}^2 + \omega_{O_z}^2 = 1 \tag{18}$$

Now, considering the time derivative of the matrix ${}^fR_m$, the angular velocity of the mobile platform is given by:

$$\vec{\omega}_m = \begin{bmatrix} \omega_x \\ \omega_y \\ \omega_z \end{bmatrix} = \begin{bmatrix} \sin\theta \, \dot{\psi} \\ \dot{\theta} \\ \cos\theta \, \dot{\psi} \end{bmatrix} \tag{19}$$

From (19) a relationship between the components $\omega_x$ and $\omega_z$ can be established. For a $\hat{\$}_O$ this relationship can be written as:

$$\omega_{O_x} = \frac{\sin\theta}{\cos\theta}\omega_{O_z} \tag{20}$$

The expression (20) is the fifth equation that completes the no nonlinear system to solve each $\hat{\$}_O$. In this study, there are four $\hat{\$}_O$; thus, it is possible to combine $\binom{4}{2}$, i.e., six possible $\Omega_{i,j}$ indices, that is,

$$\begin{aligned}
\Omega_{1,2} &= arccos\left(\vec{\mu}_{\omega_{O_1}} \cdot \vec{\mu}_{\omega_{O_2}}\right) & \Omega_{2,3} &= arccos\left(\vec{\mu}_{\omega_{O_2}} \cdot \vec{\mu}_{\omega_{O_3}}\right) \\
\Omega_{1,3} &= arccos\left(\vec{\mu}_{\omega_{O_1}} \cdot \vec{\mu}_{\omega_{O_3}}\right) & \Omega_{2,4} &= arccos\left(\vec{\mu}_{\omega_{O_2}} \cdot \vec{\mu}_{\omega_{O_4}}\right) \\
\Omega_{1,4} &= arccos\left(\vec{\mu}_{\omega_{O_1}} \cdot \vec{\mu}_{\omega_{O_4}}\right) & \Omega_{3,4} &= arccos\left(\vec{\mu}_{\omega_{O_3}} \cdot \vec{\mu}_{\omega_{O_4}}\right)
\end{aligned} \tag{21}$$

For the proximity to Type II singularity in the PR under study, the $\Omega_{i,j}$ (selected from (21)) that remains close to zero is analyzed, and verified by the equation corresponding to the linear components $\vec{\mu}_{v_O}^*$.

From a theoretical perspective, in the workspace of the PR under study there are cases where two or three different $\Omega_{i,j}$ are zero simultaneously. In these particular cases, three and four limbs degenerate the OTS system of the PR with $\|J_D\|$ less than $10^{-5}$. However, they are very difficult to reach experimentally because a singular configuration is already met for values higher than $10^{-5}$. Thus, they are not included in this paper.

### 3.3. Assembly mode equations

To determine the number of assembly modes of a 3UPS+RPU PR, a polynomial system established from the constraint equations must be solved. This could be done using Groebner bases [36], Bézout's elimination method [37], or another multivariate polynomial solving method [38]. Considering the following variable changes:

$$\begin{aligned}
x_1 &= x_m & C_\theta &= \frac{1-x_3^2}{1+x_3^2} & C_\psi &= \frac{1-x_4^2}{1+x_4^2} \\
x_2 &= z_m & & & & \\
& & S_\theta &= \frac{2x_3}{1+x_3^2} & S_\psi &= \frac{2x_4}{1+x_4^2}
\end{aligned}$$

After replacing (29) in (15), the polynomial system that defines the solutions of the Forward Kinematic problem is:



$$\begin{aligned}
&-x_1{}^2x_3{}^2x_4{}^2 - x_2{}^2x_3{}^2x_4{}^2 + A_4x_1x_3{}^2x_4{}^2 - x_1{}^2x_3{}^2 - x_1{}^2x_4{}^2 - x_2{}^2x_3{}^2 - x_2{}^2x_4{}^2 - A_5x_2x_3x_4{}^2 \\
&\quad + A_2x_3{}^2x_4{}^2 + A_3x_1x_3{}^2 + A_3x_1x_4{}^2 - x_1{}^2 - x_2{}^2 + A_5x_2x_3 + A_1x_3{}^2 + A_1x_4{}^2 + A_4x_1 \\
&\quad + A_2 = 0
\end{aligned}$$

$$\begin{aligned}
&-x_1{}^2x_3{}^2x_4{}^2 - x_2{}^2x_3{}^2x_4{}^2 + B_7x_1x_3{}^2x_4{}^2 - x_1{}^2x_3{}^2 - x_1{}^2x_4{}^2 - B_{10}x_1x_3{}^2x_4 - x_2{}^2x_3{}^2 - x_2{}^2x_4{}^2 \\
&\quad - B_9x_2x_3x_4{}^2 + B_3x_3{}^2x_4{}^2 + B_5x_1x_3{}^2 + B_5x_1x_4{}^2 - 2B_{10}x_2x_3x_4 + B_6x_3{}^2x_4 - x_1{}^2 \\
&\quad + B_{10}x_1x_4 - x_2{}^2 + B_9x_2x_3 + B_1x_3{}^2 + B_2x_4{}^2 + B_7x_1 + B_8x_4 + B_4 = 0
\end{aligned} \quad (22)$$

$$\begin{aligned}
&-x_1{}^2x_3{}^2x_4{}^2 - x_2{}^2x_3{}^2x_4{}^2 + C_7x_1x_3{}^2x_4{}^2 - x_1{}^2x_3{}^2 - x_1{}^2x_4{}^2 - C_{10}x_1x_3{}^2x_4 - x_2{}^2x_3{}^2 - x_2{}^2x_4{}^2 \\
&\quad - C_9x_2x_3x_4{}^2 + C_3x_3{}^2x_4{}^2 + C_5x_1x_3{}^2 + C_5x_1x_4{}^2 - 2C_{10}x_2x_3x_4 + C_6x_3{}^2x_4 - x_1{}^2 \\
&\quad + C_{10}x_1x_4 - x_2{}^2 + C_9x_2x_3 + C_1x_3{}^2 + C_2x_4{}^2 + C_7x_1 + C_8x_4 + C_4 = 0
\end{aligned}$$

$$-x_1{}^2 - x_2{}^2 + D_2x_1 + D_1 = 0$$

where the coefficients of the polynomial system $A_i$, $B_i$, $C_i$ and $D_i$ are obtained by grouping the geometric terms of the robot and its active variables. It is computationally complex to solve the system of equations (22) using Groebner bases. However, this system can be solved by choosing a defined configuration of the PR to apply Groebner bases and solve it numerically. Subsequently, a maximum of 44 solutions were found for the Forward Kinematic problem.

### 3.4. Test trajectories

The 3UPS+RPU PR was designed to perform human knee rehabilitation and diagnosis, this task requires three fundamental movements [39] i) flexion of hip ii) flexion-extension of knee and iii) internal-external rotation knee. This study combines these three knee movements and ankle rotation through nine trajectories to analyze the angles $\Omega_{i,j}$ as indices to detect proximity to Type II singularities. The nine trajectories were designed to make complete use of the workspace of the PR under study. Table 2 describes the movement performed by each trajectory in the lower limb and their corresponding motions in the PR under study.

Table 2: Description of test trajectories.

|  | **Lower limb** | **PR** |
|---|---|---|
| **TT1** | Complete internal-external knee rotation | Rotation around the $Z_M$ axis for the interval $\psi = [0 \quad 59]°$, with $X_F = -0.155$m |
| **TT2** | Flexion-extension of knee combined with hip flexion, ankle, and knee rotations | Simultaneous motion of 0.2m on the $X_F$ axis, 0.1m on $Z_F$, while $\theta$ rotates 15° and $\psi$ rotates 59° |
| **TT3** | Flexion-extension of knee combined with hip flexion, ankle, and knee rotations | Elliptical motion on the $Z_F$ axis as a function of 0.2m displacement on the $X_F$ axis, simultaneously turns 4° in $\theta$ and 4° for $\psi$ |
| **TT4** | Flexion-extension of knee combined with hip flexion, ankle, and knee rotations | Elliptical motion on $Z_F$ axis as a function of 0.2m displacement on the $X_F$ axis, simultaneously rotates 5° in $\theta$ and 10° for $\psi$ |
| **TT5** | Partial internal-external knee rotation | Rotation around the $Z_M$ axis for the interval $\psi = [0 \quad 20]°$, with $X_F = 0.012$m |
| **TT6** | Flexion-extension of knee combined with ankle and knee rotations | Displacement of 0.1m on the $X_F$ axis, while $\theta$ rotates 20° and $\psi$ rotates 10° |
| **TT7** | Flexion-extension of knee combined with hip flexion | Elliptical motion on $Z_F$ axis as a function of the 0.1m displacement on the $X_F$ axis |
| **TT8** | Flexion-extension of knee combined with ankle and knee rotations | Displacement of 0.16m on $X_F$ axis, while $\theta$ rotates 20° and $\psi$ rotates 20° |
| **TT9** | Flexion-extension of knee combined with ankle and knee rotations | Displacement of 0.2m on $X_F$ axis, while $\theta$ rotates 20° and $\psi$ rotates 20° |

For every test performed on the 3U<u>P</u>S+R<u>P</u>U PR the $\Omega_{3,4}$ is the closest to zero, providing the most useful information in contrast to the others five angles. The simulation of TT1 shows that $\|J_D\|$ does not reach zero. In this trajectory the minimum value of the $\Omega_{3,4}$ index (0.0193°) is reached in configuration 81 (see Fig. 7a), where the $\|J_D\|$ has a value of 0.0143 (Fig. 6a). Section 4.2 shows experimentally that for TT1 the PR loses control on its mobile platform in practice. The time between two different configurations is 0.1s for all the trajectories in this study.

In trajectories TT2-TT4 when the PR undergoes a configuration with $\|J_D\| = 0$ (see Fig. 6a, Fig. 6b), the $\Omega_{3,4}$ index is zero (see Fig. 7a, Fig. 7b). These configurations in TT2, TT3, TT4 are 157, 244 and 233, respectively. Subsequently, in the simulations of TT5-TT9 the minimum $\|J_D\|$ (see Fig. 6b, Fig. 6c) is different from zero, as is the $\Omega_{3,4}$ index (see



Fig. 7b, Fig. 7c). For TT5-TT9 the configurations mentioned are 191, 150, 335, 150 and 150, respectively. At every singular configuration detected by $\Omega_{3,4}$, it is verified that $\vec{\mu}_{v_{O_3}}^* = \vec{\mu}_{v_{O_4}}^*$.

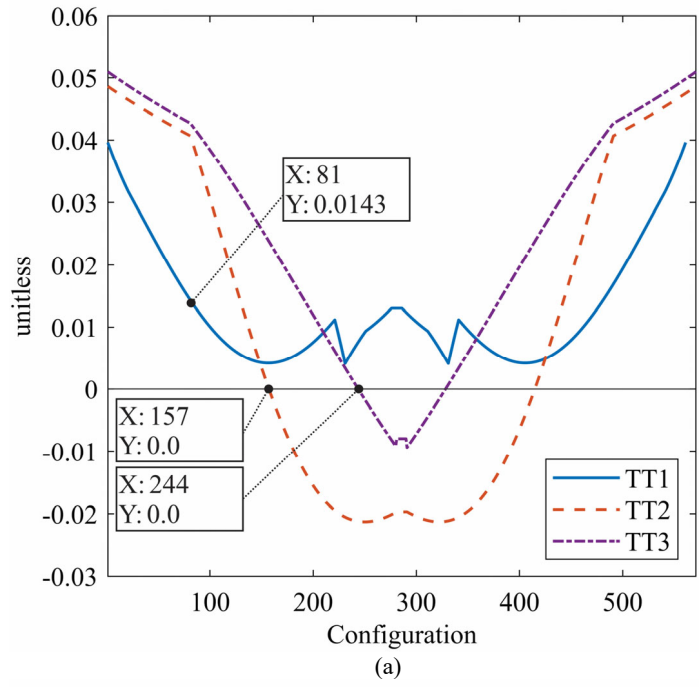

(a)

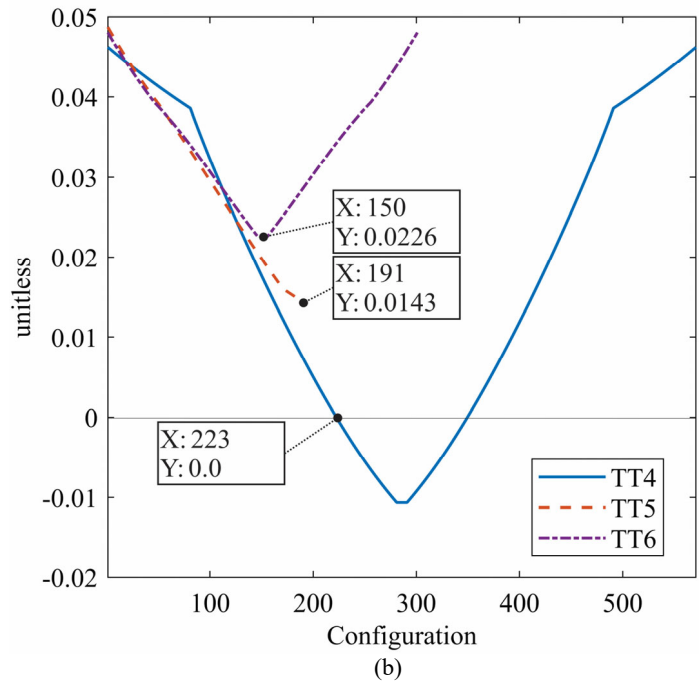

(b)



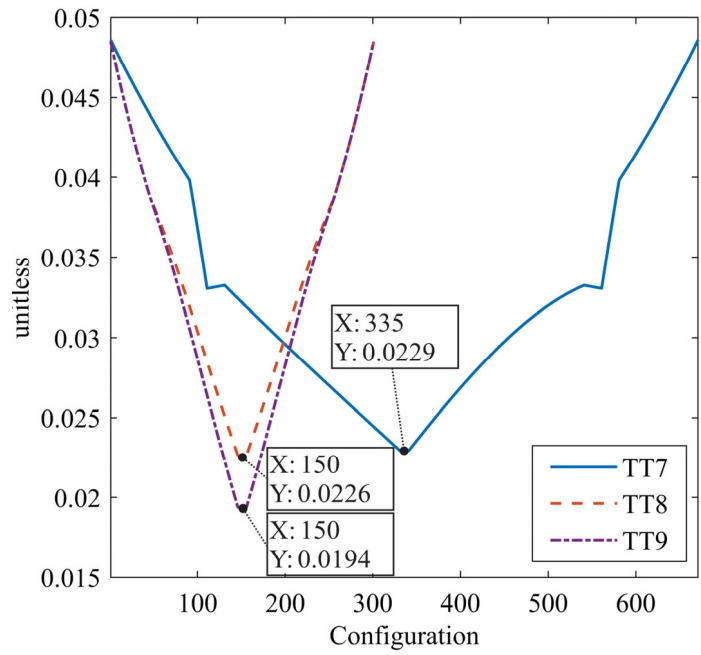

(c)

Fig. 6. $\|J_D\|$ for (a) TT1-TT3, (b) TT4-TT6, (c) TT7-TT9.

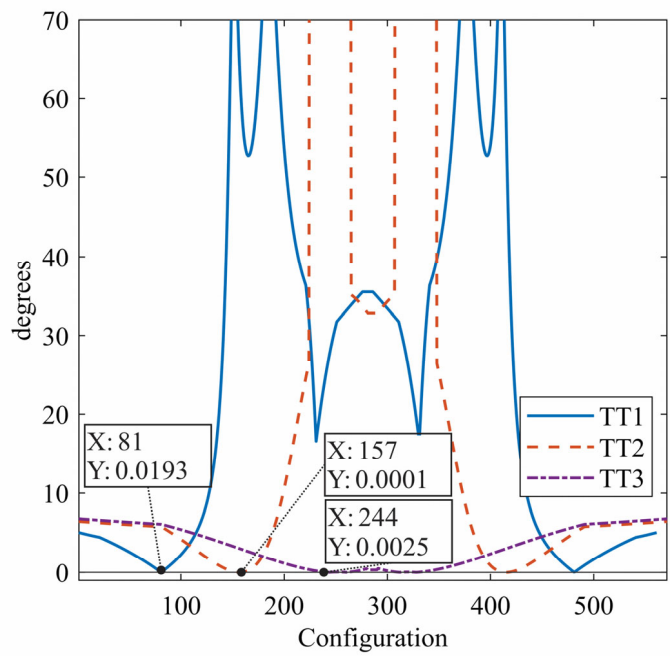

(a)



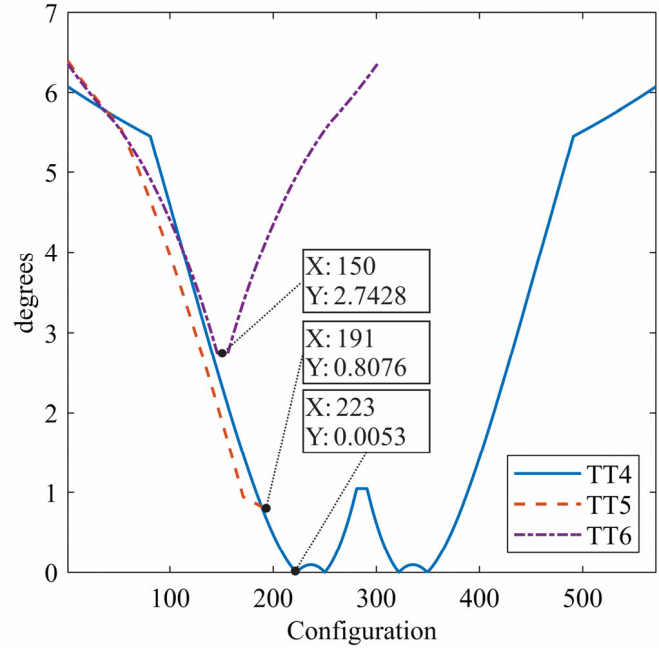

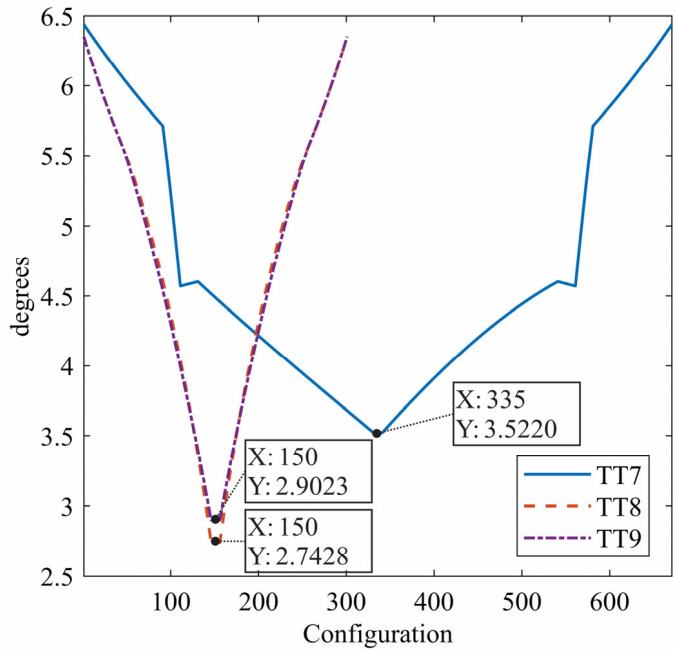

Fig. 7. $\Omega_{3,4}$ for (a) TT1-TT3, (b) TT4-TT6, (c) TT7-TT9.

All $\Omega_{i,j}$ presented in (21) are verified. Nevertheless only $\Omega_{3,4}$ is presented in Fig. 7, as it corresponds to the angle closer to the singularity than the others. Thus, in this particular trajectory, the index $\Omega_{3,4}$ is the first to detect the Type II singularities. For TT3, Fig. 8 shows that both $\Omega_{1,2}$ and $\Omega_{3,4}$ decrease until they become zero in a Type II singular configuration and $\Omega_{3,4}$ is closer to zero. In configuration 244 the $\|J_D\| = \Omega_{1,2} = \Omega_{3,4} = 0$ and verifies that $\vec{\mu}_{v_{O_1}}^* = \vec{\mu}_{v_{O_3}}^* = \vec{\mu}_{v_{O_4}}^*$. According to these simulation results, we can establish that the $\Omega_{i,j}$ closest to zero is the most suitable for detecting the proximity to Type II singularities, even though any $\Omega_{i,j}$ can detect a Type II singularity. It is important to mention that the index $\Omega_{i,j}$ and vectors $\vec{\mu}_{v_{O_i}}^*, \vec{\mu}_{v_{O_j}}^*$ are based on the linearly independent property between the $\hat{\$}_O$ previously demonstrated by Wang et al. [14].



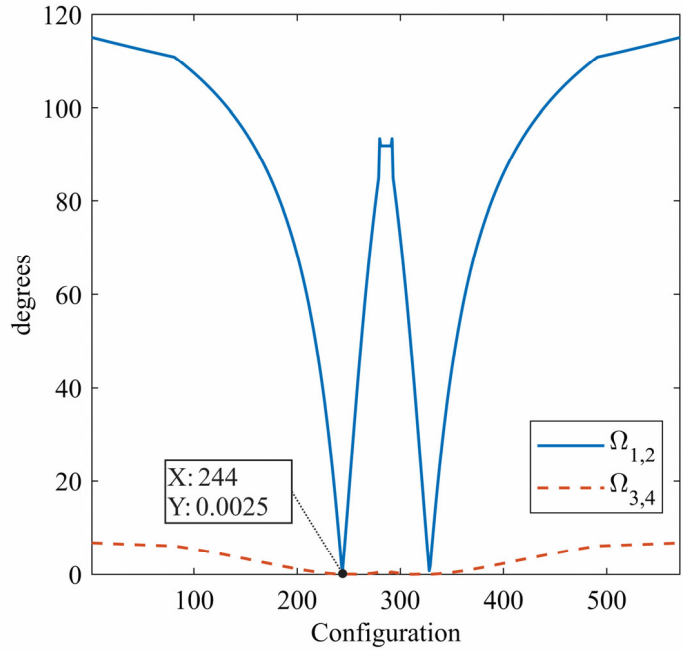

Fig. 8. Two different $\Omega_{i,j}$ indices for TT3

## 3.5. Verification trajectories

The verification trajectories differ from knee movements and have the feature of starting with $\|J_D\| \gg 0$, i.e. a non-singular configuration, and during the trajectories the $\|J_D\|$ decreases up to the final position. The description of the movements of the three trajectories is shown in Table 3.

Table 3: Description of verification trajectories.

| | |
|---|---|
| **VT1** | Independent linear motions to reach $\boldsymbol{X_F = 0.2174}$m, $\boldsymbol{Z_F = 0.7052}$m, $\boldsymbol{\theta = 27.74°}$ and $\boldsymbol{\psi = 14°}$ |
| **VT2** | Independent linear motions to reach $X_F = 0.087$m, $Z_F = 0.705$m, $\theta = -3.93°$ and $\psi = 3.38°$ |
| **VT3** | Independent linear motions to reach $X_F = 0.088$m, $Z_F = 0.724$m, $\theta = 6.39°$ and $\psi = 15.66°$ |

The decreasing behavior of the verification trajectories for the $\|J_D\|$ and the $\Omega_{3,4}$ index are shown in Fig. 9a and Fig. 9b, respectively.

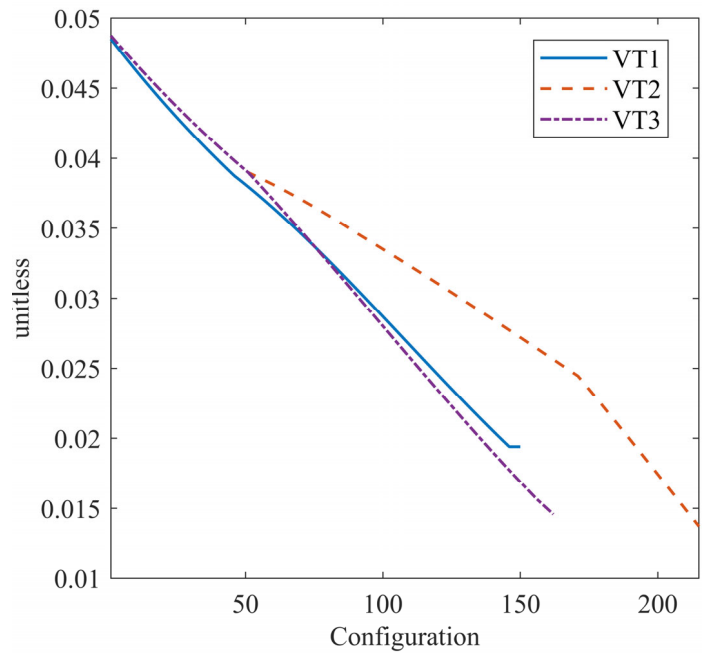



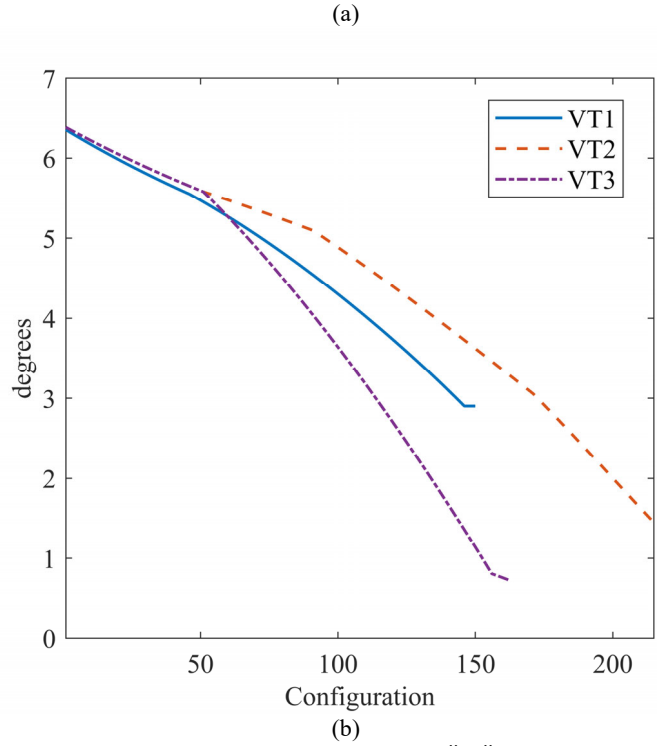

(a)

(b)

Fig. 9. Verification trajectories (a) $\|J_D\|$, (b) $\Omega_{3,4}$.

In this Section, the verification trajectories are used to analyze the sensitivity of the $\|J_D\|$ and the $\Omega_{3,4}$ index in the proximities of a Type II singularity. For this purpose, the average rate of change of both indices will be analyzed. However, the $\|J_D\|$ and $\Omega_{3,4}$ have different dimensions, so it is necessary to normalize the $\|J_D\|$ and $\Omega_{3,4}$ by dividing them by their respective maximum value. The normalized indices are represented by $\|J_D\|_n$ and $\Omega_{3,4_n}$, respectively. For the verification trajectories, the rate of change of $\|J_D\|_n$ and $\Omega_{3,4_n}$ considering the initial values as the maximum can be written as:

$$rate\ of\ change = \frac{d\|J_D\|_n}{dt} = \frac{d\Omega_{3,4_n}}{dt} \tag{23}$$

Fig. 10 shows the rate of change for each verification trajectory for the $\|J_D\|$ and the $\Omega_{3,4}$ index. In this figure, the $\|J_D\|$ and the $\Omega_{3,4}$ index are represented by a continuous and dashed line, respectively. Fig. 10a and Fig. 10c show that the $\Omega_{3,4}$ index decreases $0.1\ \%/s$ faster than the $\|J_D\|$. In Fig. 10b the $\Omega_{3,4}$ index decreases $0.05\ \%/s$ faster than the $\|J_D\|$. Both cases allow us to assume that the $\Omega_{3,4}$ index, which is the closest to zero, is more sensitive in the proximity of a Type II singularity than the $\|J_D\|$. In Section 4.3 this fact is also verified experimentally.

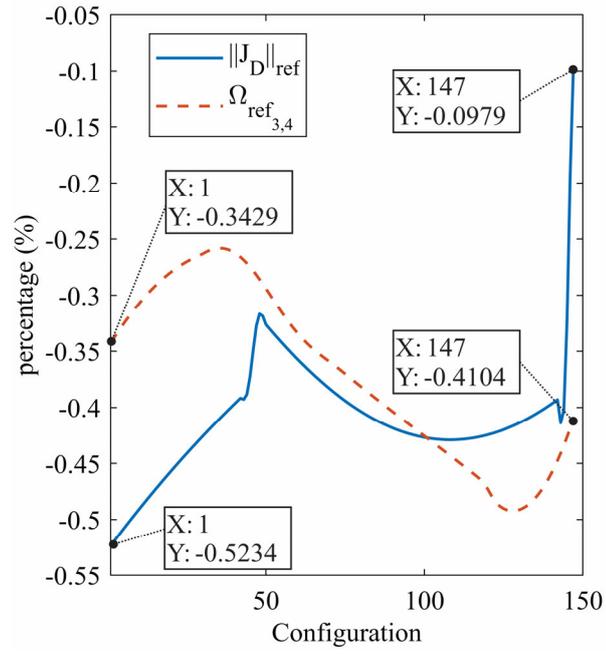

(a)

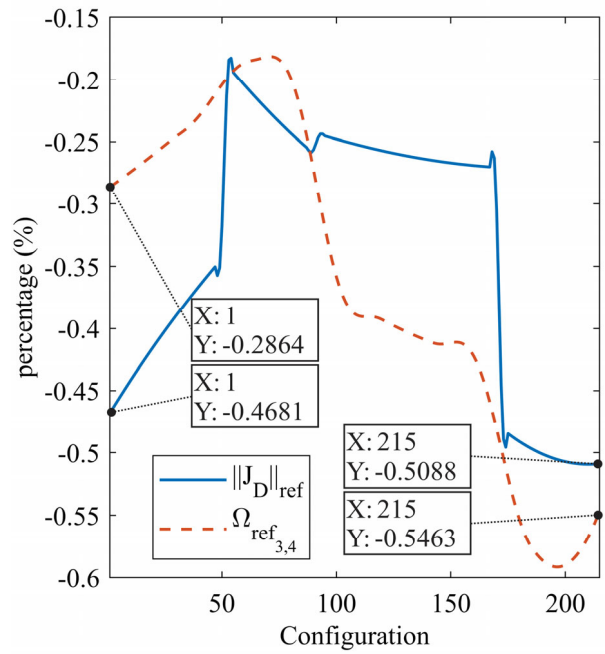

(b)

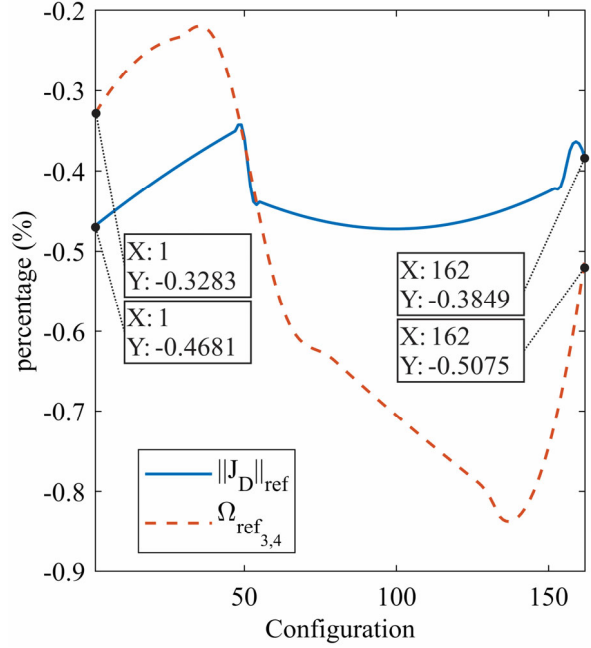

(c)

Fig. 10. Comparison of rate of change between $\|J_D\|$ and $\Omega$ (a) VT1, (b) VT2, (c) VT3.

### 3.6. Trajectories with assembly change points (AC points)

The trajectories presented in this Section have a final configuration with the $\Omega_{3,4}$ index equal to zero; however, $\|J_D\|$ is greater than zero and $\vec{\mu}^*_{v_{O_3}} \neq \vec{\mu}^*_{v_{O_4}}$. Table 4 provides a brief description of the movements performed in each trajectory.

Table 4: Description of assembly change trajectories.

| | |
|---|---|
| ACT1 | Independent linear motions to reach $X_F = 0.016$m, $Z_F = 0.7076$m, $\theta = -14.67°$ and $\psi = 20°$ |
| ACT2 | Independent linear motions to reach $X_F = -0.1$m, $Z_F = 0.75$m, $\theta = -15°$ and $\psi = 0°$ |
| ACT3 | Independent linear motions to reach $X_F = -0.144$m, $Z_F = 0.7047$m, $\theta = 7.78°$ and $\psi = 16.8°$ |

From definition (7) and the assumptions presented in Section 3.4, any $\Omega_{i,j}$ index is expected to be zero only when a Type II singularity is present. However, after applying the $\Omega_{3,4}$ index to points in the PR's workspace, configurations were found with $\|J_D\| \neq 0$, $\vec{\mu}^*_{v_{O_3}} \neq \vec{\mu}^*_{v_{O_4}}$ and $\Omega_{3,4} = 0$. Having a $\|J_D\| \neq 0$ and $\vec{\mu}^*_{v_{O_3}} \neq \vec{\mu}^*_{v_{O_4}}$, these configurations cannot be considered as Type II singularities or cusp points. With $\vec{\mu}^*_{v_{O_i}} \neq \vec{\mu}^*_{v_{O_j}}$ and $\Omega_{i,j} = 0$ a partial degeneration of two $\hat{\$}_O$ is detected. Under these conditions the PR could lose control of at least one DOF because two actuators are contributing to the angular motion of the PR in the same direction.

The simulations of ACT1 and ACT2 show the final configuration with the $\|J_D\|$ higher than zero (Fig. 11a) and the $\Omega_{3,4}$ index equal to zero (Fig. 11b). At the same configurations $\vec{\mu}^*_{v_{O_3}} \neq \vec{\mu}^*_{v_{O_4}}$, this result verifies that they are AC points. Moreover, Fig. 11 shows the advantage of $\Omega_{3,4}$ index, compared to the use of $\|J_D\|$, in AC points detection.

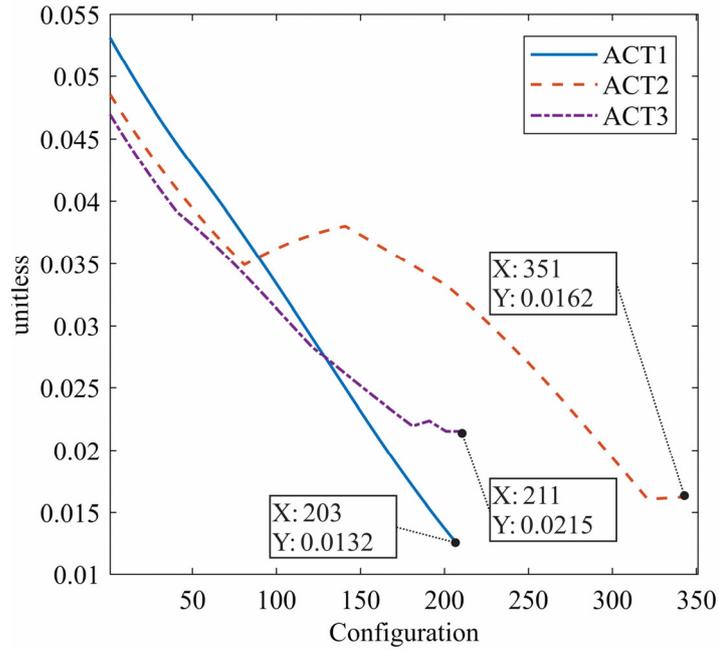

(a)

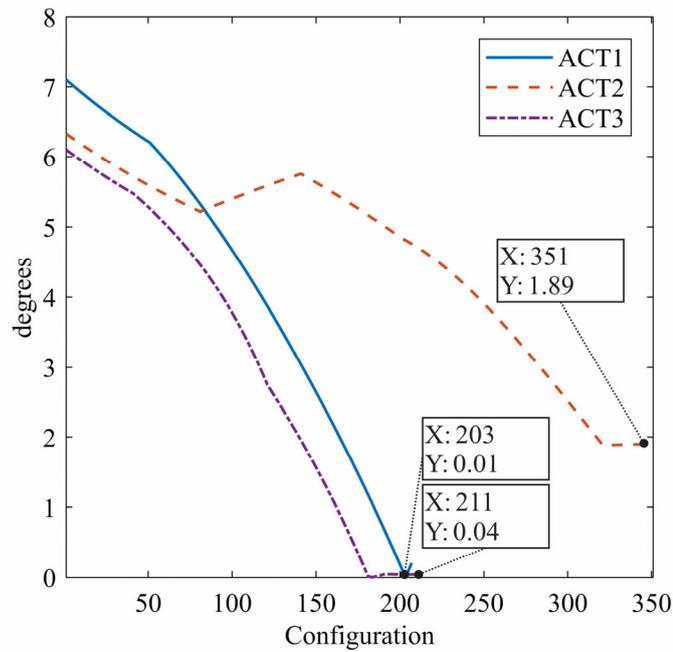

(b)

Fig. 11. Assembly change trajectories (a) $\|J_D\|$ and (b) $\Omega_{3,4}$.

In ATC3 the final configuration has the $\|J_D\|$ greater than zero (Fig. 11a) and the $\Omega$ index equal to 1.89º (Fig. 11b). In simulation, this final point is not an AC point, but in Subsection 4.4 the actual PR reaches an AC point, allowing an assembly change to be performed.

## 4. Experimental Results

At the beginning of this Section, the required experimental methodology is described. The experimental methodology focuses on the actual 3U$\underline{P}$S+R$\underline{P}$U PR, its unit control, and the high-precision photogrammetry system (5 tenths of a millimeter).

Subsequently, the three set of trajectories previously described in Section 3 are performed. The performance of the trajectories has specific objectives that aim to fulfill the three contributions of this study in the following ways:

- Test Trajectories: First, it is verified that a Type II singularity is presented with values of $\|J_D\| \approx 0$. Next, it allows the experimental benchmark to be established between the $\|J_D\|$ and the $\Omega_{i,j}$ index. Finally, the experimental benchmark established is applied to set limits for both $\|J_D\|$ and $\Omega_{i,j}$ that avoid the actual robot reaching a Type II singularity.
- Verification Trajectories: It is confirmed that the experimental limits for both $\|J_D\|$ and the $\Omega_{i,j}$ index make it possible to avoid the Type II singular zone. Moreover, these trajectories show that the $\Omega_{i,j}$ index decreases faster than $\|J_D\|$ in the proximities of a Type II singularity.
- Trajectories with assembly change points: First, it is shown that in the AC points at the end of these trajectories the actual PR loses control of the mobile platform's movements. Subsequently, an external force is applied to the mobile platform while the actual PR is at the AC point. The external force makes the actual PR move to another configuration without moving the actuators' position. Finally, this uncontrolled movement is compared with the singular and non-singular assembly change.

*4.1. Experimental methodology*

In this section, the trajectories established in Subsections 3.4 - 3.6 are performed on the actual 3U$\underline{P}$S+R$\underline{P}$U PR. The reference trajectories ($\vec{X}_{ref}$) are transformed to the space of generalized active coordinates $\vec{q}_{ind}$ by solving the Inverse kinematics problem (see Fig. 12a). These $\vec{q}_{ind}$ trajectories are used as a reference for the robot control unit. This control unit is embedded in a high-performance industrial PC. It is equipped with data acquisition cards for reading the encoders' signal and supplying the control actions for the PR actuators (active generalized coordinates $q_{13}$, $q_{23}$, $q_{33}$ and $q_{42}$). Next, the measure of the position and orientation of the mobile platform ($\vec{X}'_{med}$) is smoothed before being analyzed. Signal smoothing is performed in order to i) reduce the noise on the measurements and ii) adjust the sampling time from 8.3ms to 0.1s. The signal smoothing uses the Loess algorithm by Cleveland [40]. After performing a trajectory, the reference $\vec{X}_{ref}$ and the filtered position and orientation of the mobile platform ($\vec{X}_{med}$) are stored. The experimental analysis of singularities uses $\vec{X}_{ref}$ and $\vec{X}_{med}$ to calculated the $\|J_D\|$ and $\Omega_{i,j}$ and finally compare the reference and measured results.

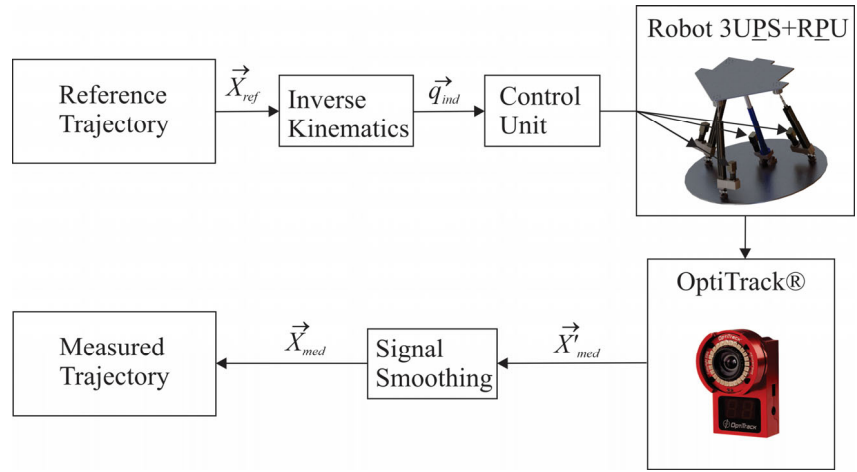

(a)

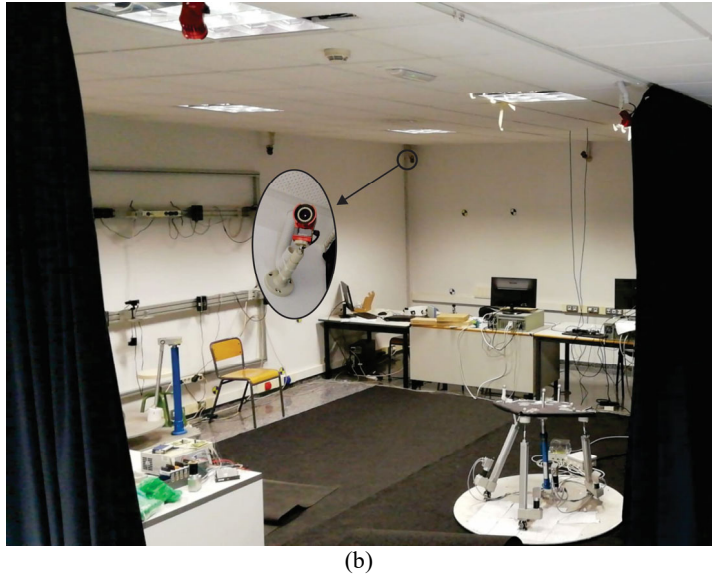
(b)
Fig. 12. Experimentation (a) diagram (b) equipment.

A critical element in the experimental study is the precise measurement of the position and orientation of the mobile platform ($\vec{X}$). For this purpose, an OptiTrack® photogrammetry system consisting of 10 infrared cameras is used (Fig. 12b). This photogrammetry system has been selected due to the following advantages. i) easy calibration of cameras that provides an accuracy of 0.5 mm. ii) sample time of 8.3 ms compatible with the controller sample time of 10 ms. iii) rigid body recognition with minimum modification of their dynamic properties. A rigid body is defined by plastic spheres (infrared markers) and iv) TCP/IP Communication protocol with .NET technology, compatible with several software applications such as Matlab and LabVIEW.

### 4.2. Test trajectories

When executing the trajectories of Table 2 on the actual PR, the trajectory TT1 was stopped in configuration 89 because of the loss of control of the mobile platform. Trajectory TT1 is theoretically non-singular. In configuration 89, where control of the PR is lost, $\|J_D\|$ is equal to 0.0148 (Fig. 13a), but not zero. In TT2, TT3, TT4 and TT5 control of the PR is lost in configurations 150, 226, 166, and 191, respectively. In those configurations the value of the $\|J_D\|$ is greater than zero (see Fig. 13a and Fig. 13b).

It was verified that the actual 3U$\underline{P}$S+R$\underline{P}$U PR had all its components correctly assembled and fixed, ruling out failures of this type. At the non-singular starting point of TT1, clearances at the actual PR joints were visually verified (on the mobile platform). The clearances were mainly presented at the spherical joints. Note that the spherical joints were fabricated by the authors, using an iron sphere and a nylon circular case. The photogrammetry system allows us to measure a maximum displacement of 2 mm and a maximum rotation of 2.5º from the origin of the mobile platform with the actuators locked. Using this experimental analysis, a direct relation between joint clearances and singularities cannot be established. However, the photogrammetry measurements implicitly consider manufacturing errors such as joint clearances. Moreover, the measurements allow to verify experimentally that a Type II singularity occurs in the vicinity of $\|J_D\| = 0$. Fig. 14a and Fig. 14b show the same results for the values of the $\Omega_{3,4}$ index.

In the case of TT6-TT9, the trajectories are performed easily. The values reached by the actual PR for the $\|J_D\|$ in each configuration are presented in Fig. 13b and Fig. 13c. The behavior of the $\Omega_{3,4}$ index in each configuration is presented in Fig. 14b and Fig. 14c. The figures related to the $\|J_D\|$ show the difficulty of knowing how far the actual PR is from the Type II singularity due to its dimensionless nature. On the other hand, the figures of the $\Omega_{3,4}$ index, due to their physical meaning, allow us to interpret the proximity to a Type II singularity. In addition, Fig. 14a, Fig. 14b and Fig. 14c show that the proposed index is able to identify that the Type II singularities are produced by the actuators on limbs three and four. For trajectories TT6-TT9 the minimum values for $\|J_D\|$ and $\Omega_{3,4}$ are presented in Fig. 13c and Fig. 14c respectively.

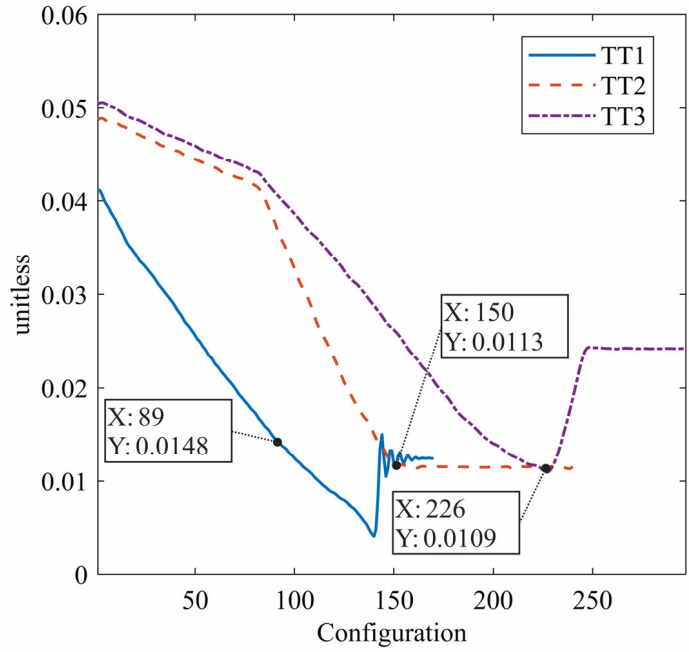

(a)

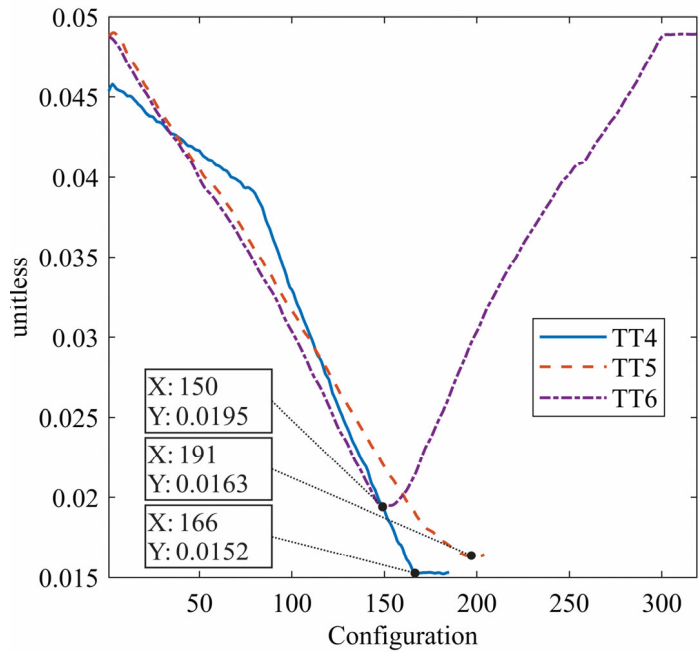

(b)

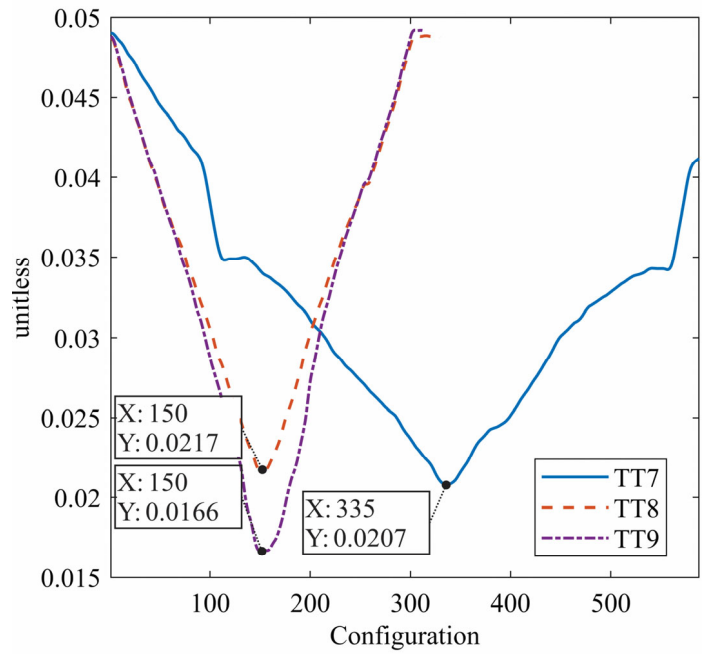

(c)

Fig. 13. Results of $\|J_D\|$ for (a) TT1-TT3, (b) TT4-TT6, (c) TT7-TT9.

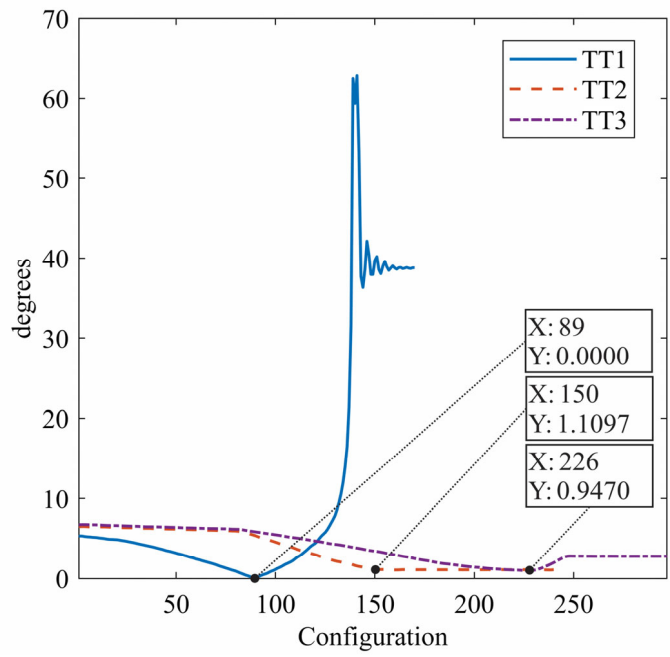

(a)

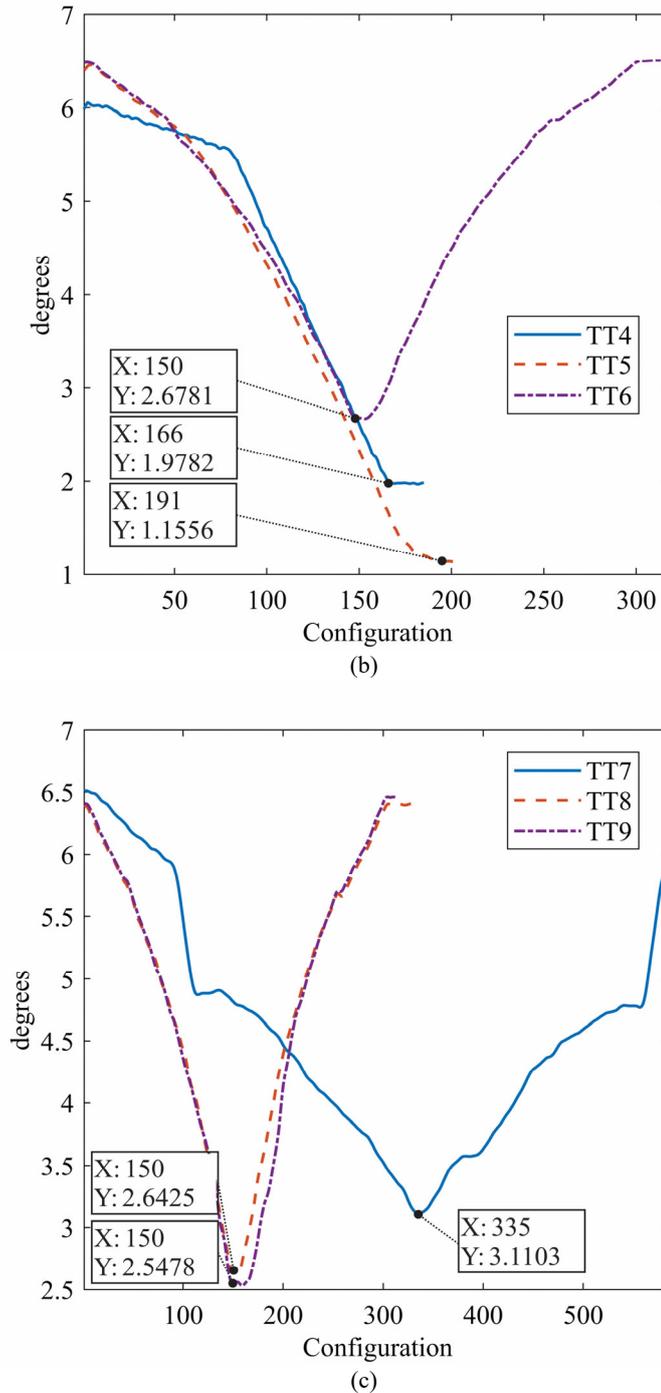

Fig. 14. Results of $\Omega_{3,4}$ for (a) TT1-TT3, (b) TT4-TT6, (c) TT7-TT9.

To ensure proper detection of a Type II singularity in an actual PR, this study proposes the experimental benchmark shown in Fig. 15. First, based on the task, the fundamental trajectories of the PR under study are selected. Subsequently, several trajectories similar to the fundamental trajectories are generated to make complete use of the workspace, especially near to theoretical Type II singularities. The number of test trajectories is represented by $n_{TT}$. Then, each test trajectory is executed on the actual PR to measure the minimum values reached for the index under analysis ($\iota_a$). In a singular trajectory the minimum value is the value reached before losing the control of the PR. The experimental limit ($lim_e$) for the $\iota_a$ is calculated as the average of the minimum values reached in the test trajectories. Finally, a new set of trajectories is developed to verify that the experimental limit can detect the proximity to a singular configuration. The number of these verification trajectories has to be at least 30% of the $n_{TT}$ and the final point has to be next to the experimental limit of $\iota_a$. If $\min(\iota_a) \geq lim_e$ in the execution of the verification trajectories does not cause control problems, the process is finished. Otherwise, it is necessary to modify the criteria to generate the test trajectories, according to the PR application. As mentioned above, the aim of the PR under study is knee and diagnosis rehabilitation, where three fundamental trajectories are the major tasks performed. After modifying these fundamental trajectories, $n_{TT} = 9$ test trajectories were designed (TT1-TT9). Thus, three verification trajectories (VT1-VT3) were required in this study. For the 3U$\underline{P}$S+R$\underline{P}$U PR the $\iota_a$ was the $\|J_D\|$ or the $\Omega_{i,j}$.

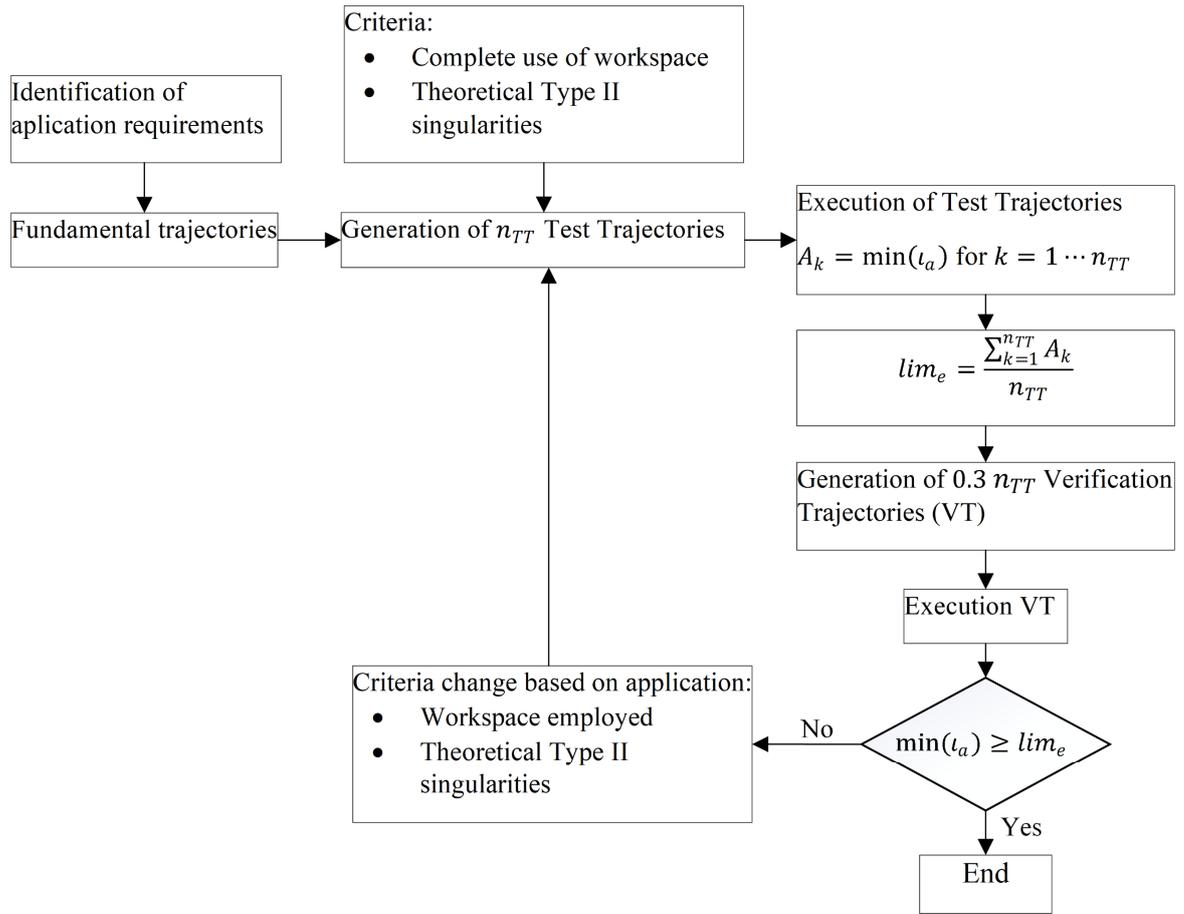

Fig. 15. Experimental benchmark flowchart.

The data used to set the experimental limits for the $\|J_D\|$ and the $\Omega_{3,4}$ index for the PR under study is shown in Table 5. The theoretical minimum values are included in this table to contrast them with the measurements on the actual PR.

Table 5: Results of minimum $\|J_D\|$ and $\Omega_{3,4}$ for test trajectories.

|  | $\|J_D\|_{min}$ (unitless) |  | $\Omega_{min}$ (°) |  | Execution |
|---|---|---|---|---|---|
|  | Theoretical | Measured | Theoretical | Measured |  |
| TT1 | 0.0041 | 0.0041 | 0.0193 | 0.0000 | It stops at sample 89 |
| TT2 | -0.0213 | 0.0113 | 0.0000 | 1.0508 | It stops at sample 150 |
| TT3 | -0.0094 | 0.0109 | 0.0011 | 0.9470 | It stops at sample 226 |
| TT4 | -0.0106 | 0.0152 | 0.0000 | 1.9782 | It stops at sample 166 |
| TT5 | 0.0143 | 0.0163 | 0.8077 | 1.1556 | It stops at sample 191 |
| TT6 | 0.0227 | 0.0195 | 2.7428 | 2.6781 | Complete |
| TT7 | 0.0229 | 0.0207 | 3.5220 | 3.1103 | Complete |
| TT8 | 0.0227 | 0.0217 | 2.7428 | 2.6425 | Complete |
| TT9 | 0.0194 | 0.0166 | 2.9023 | 2.5478 | Complete |
| Average |  | 0.0151 |  | 1.7900 |  |

Based on Table 4, the experimental limits for $\|J_D\|$ and $\Omega_{3,4}$ are obtained through averaging the minimal values reached in the singular and non-singular trajectories. The limit for the $\|J_D\|$ results in 0.015, this value is approximated to thousandths of a unit, as the $\|J_D\|$ has dimensionless values between [0 1]. For the case of the $\Omega_{3,4}$ index the limit is set to 1.80°. In this case, the analysis in thousandths does not represent an advantage because it is a physical magnitude with a range between [0 180]. Then, the approximation of the limit value for the $\Omega_{3,4}$ index is limited to hundredths of a degree.

The proposed benchmark (Fig. 15) represents the second contribution of this paper, it can be applied based on $\|J_D\|$ or $\Omega_{i,j}$ as both of them decrease in the proximity of a Type II singularity. The proposed benchmark can be applied to another PR by modifying the set of trajectories based on the requirements of the new PR application. The set of test trajectories must be carefully selected because they directly affect the experimental limit of $\Omega_{i,j}$. For the PR under study by using only nine singular and non-singular trajectories ensures enough accuracy for the specific knee rehabilitation application.

*4.3. Verification trajectories*

After performing trajectory VT1, the mobile platform does not lose control of its motions throughout the test. Fig. 16a and Fig. 16b show the values of the $\|J_D\|$ and the $\Omega_{3,4}$ index, respectively. In trajectory VT1 the minimum value of the $\|J_D\|$ and the $\Omega_{3,4}$ index are higher than the limits established in the previous Section (see Table 6). In the final configuration of trajectory VT2 the values of the $\|J_D\|$ and $\Omega_{3,4}$ index are below the established limits (see Fig. 16a and Fig. 16b, respectively). At the end of VT2 an external force is applied, so the parallel robot cannot maintain its position (Type II singularity). VT1 and VT2 confirm that the experimental limits for the $\|J_D\|$ and $\Omega$ index were correctly established. The data of the $\|J_D\|$ and the $\Omega_{3,4}$ index for both the theoretical and experimental part of the verification trajectories are depicted in Table 6.

Table 6: Results of verification trajectories.

|  | $\|J_D\|_{min}$ (su) |  | $\Omega_{min}$ (°) |  |
| --- | --- | --- | --- | --- |
|  | Theoretical | Measured | Theoretical | Measured |
| **VT1** | 0.0194 | 0.0166 | 2.90 | 2.54 |
| **VT2** | 0.0137 | 0.0140 | 1.44 | 1.59 |
| **VT3** | 0.0145 | 0.0174 | 0.73 | 1.29 |

Finally, trajectory VT3 is performed easily, where the final configuration has the $\|J_D\|$ greater than the experimental limit. However, for the same configuration the $\Omega_{3,4}$ index is less than the experimental value set (see Table 6). It will be explained in detail in the next Subsection.

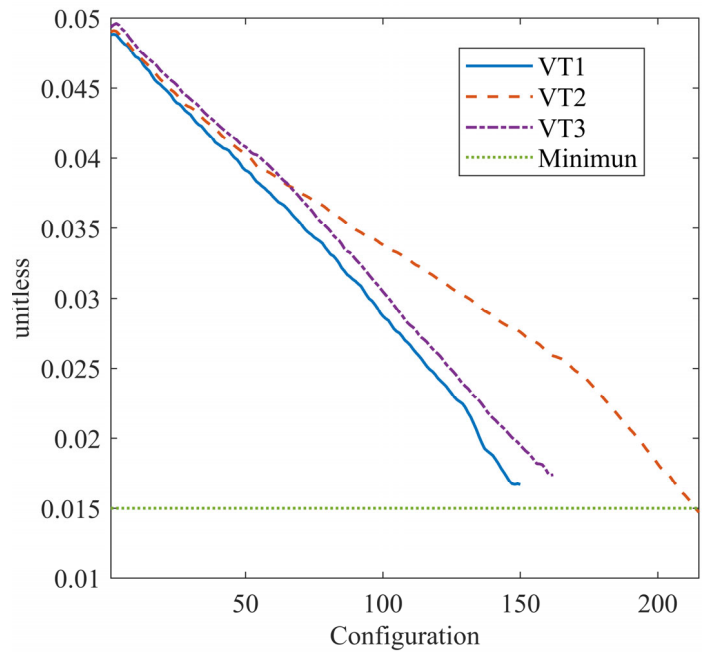

(a)

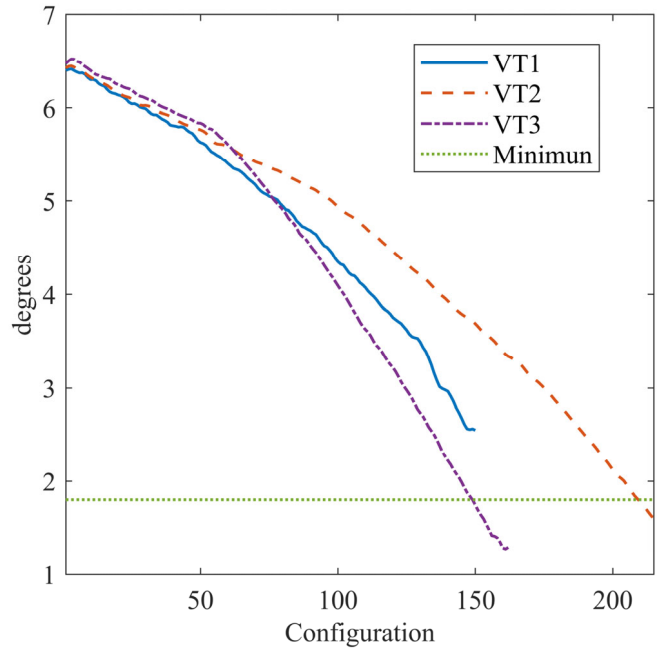

(b)

Fig. 16. Results of verification trajectories (a) $\|J_D\|$, (b) $\Omega_{3,4}$.

Fig. 17a, Fig. 17b and Fig. 17c show the rate of change between the $\|J_D\|$ and the $\Omega_{3,4}$ index for the execution of trajectories VT1, VT2 and VT3, respectively. Fig. 17 shows that the $\Omega_{3,4}$ index has an increase in the rate of change of at least $0.05\ \%/_s$. The actual data presented confirm that the $\Omega_{3,4}$ index decreases faster as it gets closer to a Type II singularity. This verifies experimentally that the $\Omega_{3,4}$ index has greater sensitivity in the detection of Type II singularity proximity.

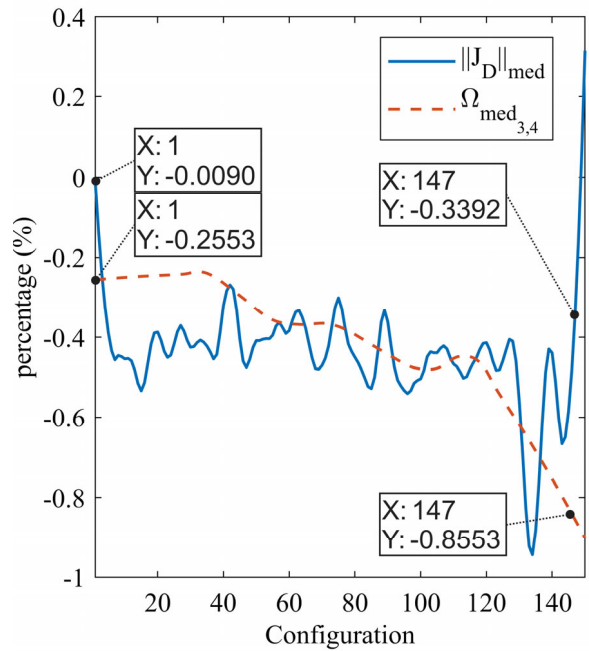

(a)

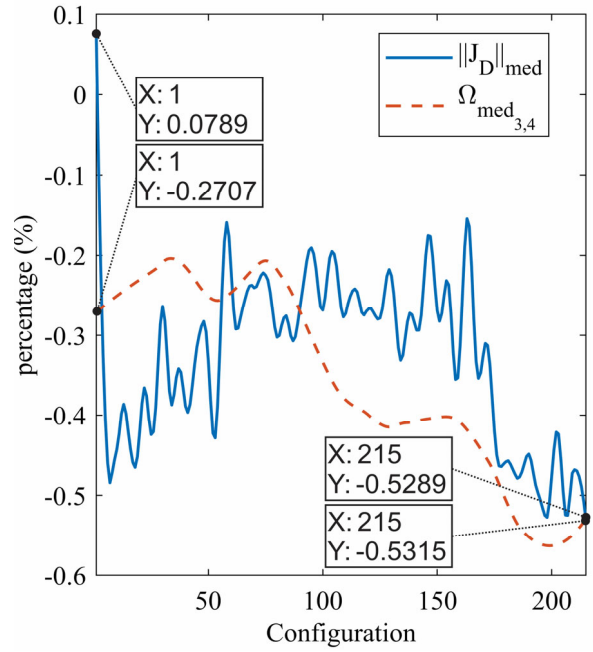

(b)

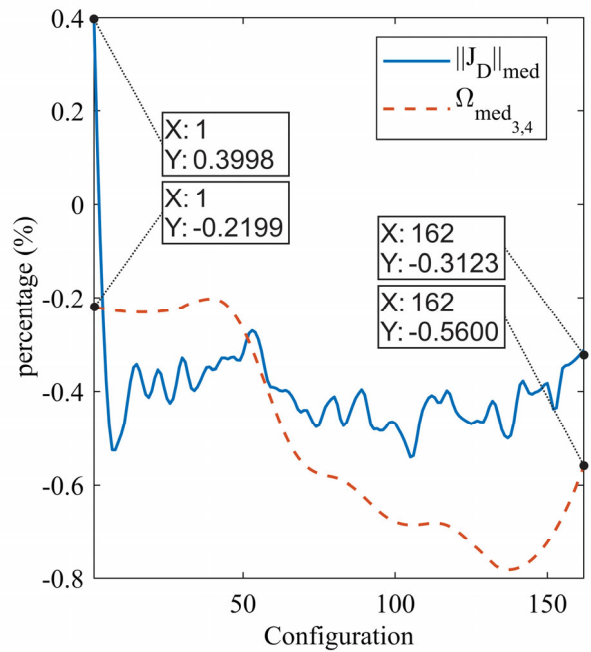

(c)

Fig. 17. Results for change velocity of $\|J_D\|$ and $\Omega_{3,4}$ for (a) VT1, (b) VT2, (c) VT3.

### 4.4. Trajectories with AC points

In the end configuration of trajectory VT3, the actual PR is not able to keep its position while an external force is applied to the mobile platform. In this configuration the $\|J_D\|$ calculated (Fig. 16a) is greater than the experimental limit but the $\Omega_{3,4}$ index is below the experimental limit (Fig. 16b). There is a partial degeneration of a pair of $\hat{\$}_O$. In this Section, three trajectories executed on the actual PR with the same characteristic in the final configuration are analyzed.

Fig. 18a and Fig. 18b show the initial and final values reached by the $\|J_D\|$ and $\Omega_{3,4}$ respectively, during the application of the external force. Note that in the trajectory shown in Fig. 18 the condition $\vec{\mu}_{vO_3}^* \neq \vec{\mu}_{vO_4}^*$ is verified.

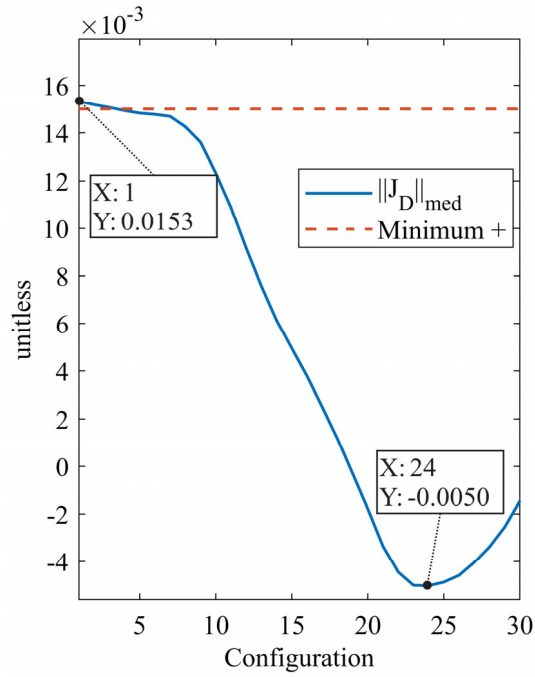

(a)

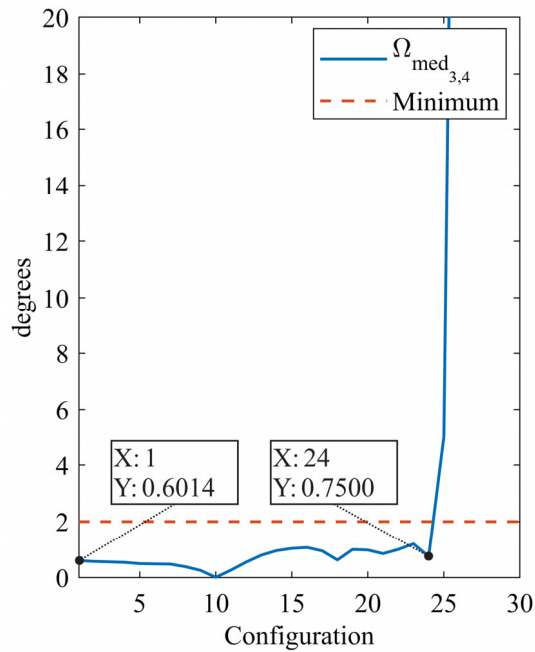

(b)

Fig. 18. Assembly change performed at the end of ATC1 (a) $\|J_D\|$ and (b) $\Omega_{3,4}$.

Based on the final configuration reached by the actual PR in ACT1, the Forward Kinematic problem is solved using the expression (22) from Subsection 3.3. It determines 44 possible positions in which the 3U$\underline{P}$S+R$\underline{P}$U PR can theoretically be assembled, 8 of which are real and the others are imaginary. From the 8 real possible assembly modes, the solutions with negative robot heights ($z_m$) are eliminated, leaving 3 theoretical assembly modes (Fig. 19).

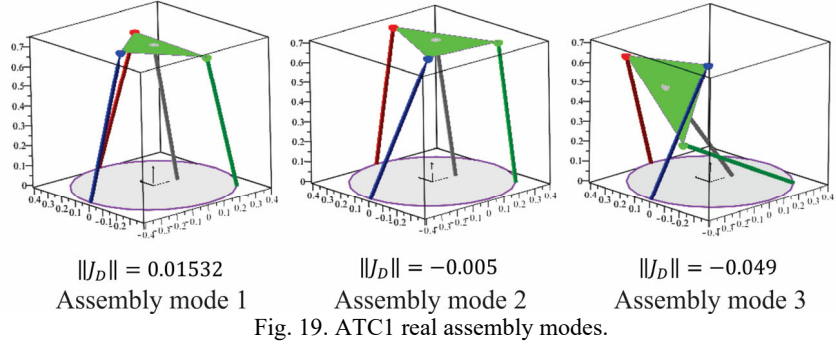

$\|J_D\| = 0.01532$  
Assembly mode 1

$\|J_D\| = -0.005$  
Assembly mode 2

$\|J_D\| = -0.049$  
Assembly mode 3

Fig. 19. ATC1 real assembly modes.

Fig. 19 shows that the $\|J_D\|$ at the initial and final points reached by the actual PR in Fig. 18a match assembly mode 1 and assembly mode 2. This verifies the existence of a singular assembly change in the AC point reached at the end of ACT1. The final configuration of trajectory ATC2 has a similar behavior to that previously analyzed. In the final configuration of ACT2 there are 8 real assembly modes that can be reached, and 4 of them can actually be assembled. In ACT2 a singular assembly change was performed by the actual PR when the external force was applied.

Finally, for the AC point at the end of ATC3, 4 real assembly modes can be achieved, two of which are feasible. Fig. 20 shows the non-singular assembly change performed by applying an external force to the mobile platform.

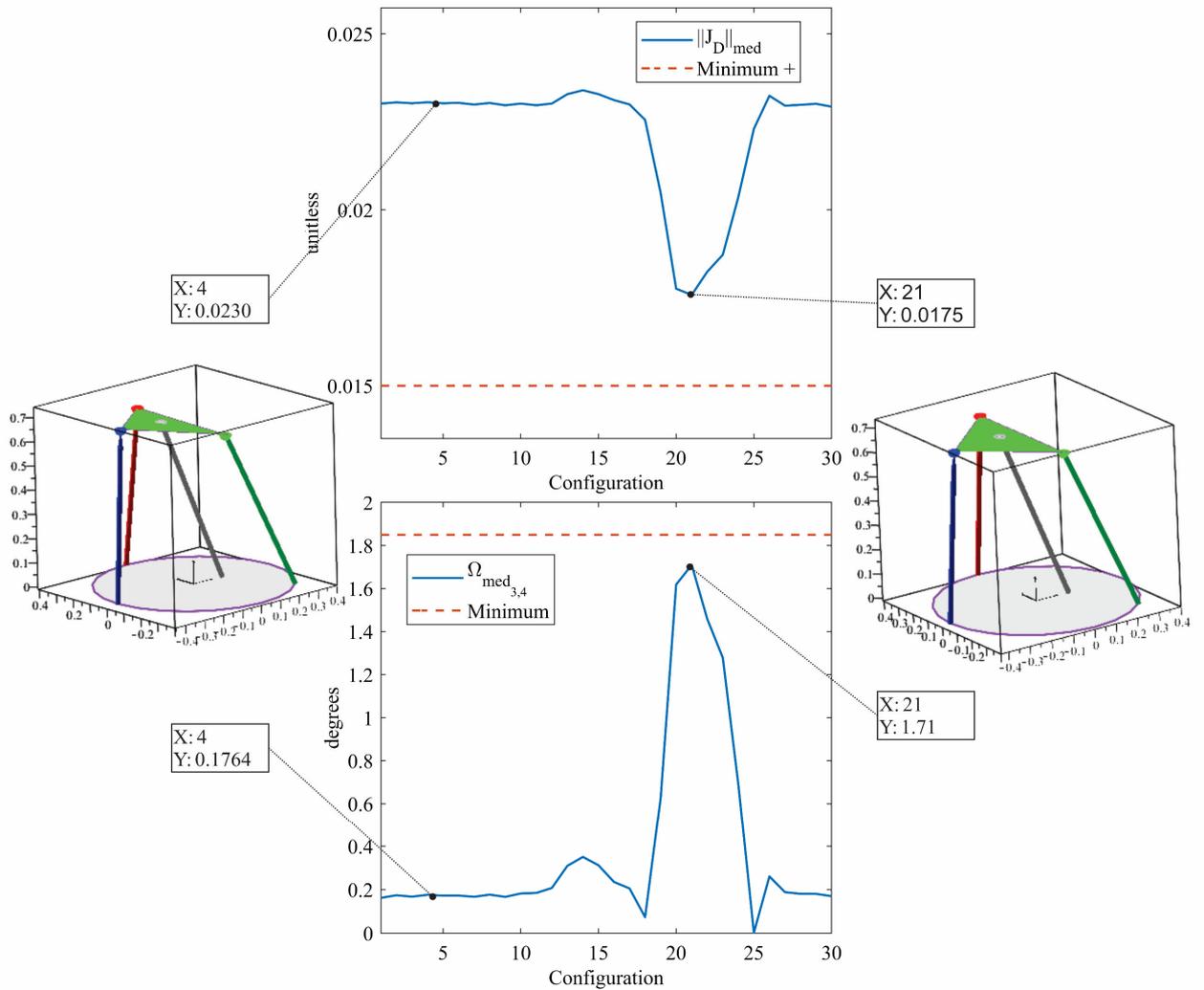

Fig. 20. $\|J_D\|$ for assembly change in ATC3, with assembly modes reached.

The final configuration of ACT1-ACT3 cannot be considered a Type II singularity. This is because the $\|J_D\|$ has a value greater than the experimental limit and the theoretical condition of $\|J_D\| = 0$ and $\overrightarrow{\mu}_{v_{O_3}}^* \neq \overrightarrow{\mu}_{v_{O_4}}^*$ is not suitable. However, the $\Omega_{3,4}$ index has a value below the established limit for a Type II singularity. The previous experimental observation allows us to state that, in fact, when a partial degeneration in a pair of $\hat{\$}_O$ occurs, the actual robot loses the control capacity of at least one DOF. A configuration under this condition is not necessarily a Type II singularity or a cuspidal point. Nevertheless, the behavior for practical purposes is equivalent, due to loss of control of the mobile platform.

The proposed indices could identify a novel non-singular configuration where an actual robot loses control of the end effector. These configurations are the third contribution of this paper. These non-singular configuration are named AC points because of the assembly change performed by the actual 3UPS+RPU PR. It is necessary to highlight the capability of a $\Omega_{i,j}$ index to detect this type of points, both analytically and experimentally, something that cannot be done by $\|J_D\|$.

## 5. Conclusions

From the results obtained in this study, it can be concluded that:

The proposed index $\Omega_{i,j}$, as a detection index for proximity to Type II singularities, has a physical meaning and a higher sensitivity near singularities. The higher sensitivity of the proposed index was verified from an analytical and experimental perspective in contrast to $\|J_D\|$. Moreover, the proposed index based on the angle between two instantaneous screw axes from the Output Twist Screws (OTSs) is capable of identifying the pair of active components causing the Type II singularity. In other words, the $\Omega_{i,j}$ index detects when two actuators are contributing to the motion of the mobile platform on a same $\hat{\$}_{O_i}$, with $\vec{\mu}_{v_{O_i}}^* = \vec{\mu}_{v_{O_j}}^*$.

The linear component $\vec{\mu}_{v_O}^*$ of two different $\hat{\$}_O$ in Type II singularities are equal, and in this study it is used only to verify Type II singularities. However, a partial degeneration has not been developed for this component; thus, further studies are required.

An experimental setup that uses a photogrammetry system in conjunction with an industrial computer-based control system to measure the position of the actual 3UPS+RPU PR was presented. The experimental setup allows to show, for the first time, that an actual Type II singularity occurs for $\|J_D\| > 0$ and $\Omega_{i,j} > 0$. Based on the experiments, a benchmark has been developed to analyze Type II singularities. The proposed experimental benchmark uses trajectories that progressively move away from a Type II singularity to establish the limit value of the $\|J_D\|$ or the $\Omega_{i,j}$ index. The experimental limits guarantee that the actual PR does not reach a singularity and avoids the need to model the error produced by clearances in the PR joints. This study was not intended to establish a direct relation between Type II singularities and manufacturing errors such as joint clearances. The research was focused on developing an experimental approach to establish the limits of $\|J_D\|$ or $\Omega_{i,j}$ to determine singular configuration on an actual PR. The experimental benchmark proposed in this paper can be applied to other types of PRs changing only the set of trajectories according to the application of the PR. The experimental limits of the $\|J_D\|$ and the $\Omega_{i,j}$ index can be applied to optimization processes in the design of a PR or in trajectory planning.

The proposed index $\Omega_{i,j}$ renders it possible to identify novel configurations of the robot where two angular motions ($\vec{\mu}_{v_{O_i}}^* \neq \vec{\mu}_{v_{O_j}}^*$) produced by a pair of actuators are aligned with $\|J_D\| \neq 0$. This particular configuration cannot be considered a Type II singularity, much less a cuspidal point. Through the experiments, it was verified that the actual PR loses control of its final effector when it reaches these particular configurations. In this configuration, the actual PR undergoes from one assembly mode to another, by the application of an external force. For this reason, we named the particular configurations as assembly change points or AC points. It is important to analyze and identify these points because of the loss of control of at least one degree of freedom of the PR presented at these points. In an AC point the behavior of the actual PR is analogous to a Type II singularity, which can be dangerous for the user or the robot itself. Moreover, it is important to mention that the behavior shown by AC points were proven just for a 3UPS+RPU PR assembled according to Table 1, and thus, further studies are required.

The developed experimental analysis for detecting Type II singularities has two limitations. First, the $\Omega_{i,j}$ index has been tested only in a limited PR with 4-DOF. Second, the generalization of the experimental benchmark has to be further verified by implementing the indices on a different parallel robot.

The 3UPS+RPU PR used in this study is being developed for knee rehabilitation. In this regard, the main contribution of this paper is that this index can be implemented for trajectory planning and design of an advanced controller, focusing on avoiding Type II singularities. Therefore, the experimental limits established for $\|J_D\|$ will be used to optimize the design of the 3UPS+RPU PR. Moreover, the proposed $\Omega_{i,j}$ index will be applied to an in-depth study about the peculiar behavior of AC points in several assembly configuration. This study will focus on the partial degeneration of the angular component of an OTS.

## Acknowledgements

This work was supported by the Spanish Government [*Integración de modelos biomecánicos en el desarrollo y operación de robots rehabilitadores reconfigurables*, DPI2017-84201-R-AR]. It was also supported in part by Escuela Politécnica Nacional of Quito, [*Control adaptativo basado en inteligencia artificial aplicado a un sistema mecatrónico fundado en un robot paralelo para la diagnosis y rehabilitación*, PIMI-1504].

**References**

[1] S. Staicu, Dynamics of Parallel Robots, Springer International Publishing, Cham, 2019. https://doi.org/10.1007/978-3-319-99522-9.

[2] S. Briot, W. Khalil, Dynamics of Parallel Robots--From Rigid Links to Flexible Elements. ISBN: 978-3-319-19787-6, 2015. https://doi.org/10.1007/978-3-319-19788-3.

[3] C. Gosselin, J. Angeles, Singularity analysis of closed-loop kinematic chains, IEEE Trans. Robot. Autom. 6 (1990) 281–290. https://doi.org/10.1109/70.56660.

[4] F.C. Park, J.W. Kim, Singularity Analysis of Closed Kinematic Chains, J. Mech. Des. 121 (1999) 32–38. https://doi.org/10.1115/1.2829426.

[5] R. di Gregorio, V. Parenti-Castelli, Mobility analysis of the 3-UPU parallel mechanism assembled for a pure translational motion, in: 1999 IEEE/ASME Int. Conf. Adv. Intell. Mechatronics (Cat. No.99TH8399), IEEE, 1999: pp. 520–525. https://doi.org/10.1109/AIM.1999.803224.

[6] M. Slavutin, O. Shai, A. Sheffer, Y. Reich, A novel criterion for singularity analysis of parallel mechanisms, Mech. Mach. Theory. 137 (2019) 459–475. https://doi.org/10.1016/j.mechmachtheory.2019.03.001.

[7] P.A. Voglewede, I. Ebert-Uphoff, Measuring "closeness" to singularities for parallel manipulators, in: IEEE Int. Conf. Robot. Autom. 2004. Proceedings. ICRA '04. 2004, IEEE, 2004: pp. 4539-4544 Vol.5. https://doi.org/10.1109/ROBOT.2004.1302433.

[8] J.P. Merlet, Jacobian, Manipulability, Condition Number, and Accuracy of Parallel Robots, J. Mech. Des. 128 (2006) 199–206. https://doi.org/10.1115/1.2121740.

[9] J. Gallardo-Alvarado, R. Rodríguez-Castro, P.J. Delossantos-Lara, Kinematics and dynamics of a 4-PRUR Schönflies parallel manipulator by means of screw theory and the principle of virtual work, Mech. Mach. Theory. 122 (2018) 347–360. https://doi.org/10.1016/j.mechmachtheory.2017.12.022.

[10] J.K. Davidson, K.H. Hunt, G.R. Pennock, Robots and Screw Theory: Applications of Kinematics and Statics to Robotics, J. Mech. Des. 126 (2004) 763. https://doi.org/10.1115/1.1775805.

[11] M.S.C. Yuan, F. Freudenstein, L.S. Woo, Kinematic Analysis of Spatial Mechanisms by Means of Screw Coordinates. Part 2—Analysis of Spatial Mechanisms, J. Eng. Ind. 93 (1971) 67. https://doi.org/10.1115/1.3427919.

[12] C. Chen, J. Angeles, Generalized transmission index and transmission quality for spatial linkages, Mech. Mach. Theory. 42 (2007) 1225–1237. https://doi.org/10.1016/j.mechmachtheory.2006.08.001.

[13] Y. Takeda, H. Funabashi, Motion Transmissibility of In-Parallel Actuated Manipulators, JSME Int. Journal. Ser. C, Dyn. Control. Robot. Des. Manuf. 38 (1995) 749–755. https://doi.org/10.1299/jsmec1993.38.749.

[14] J. Wang, C. Wu, X.-J. Liu, Performance evaluation of parallel manipulators: Motion/force transmissibility and its index, Mech. Mach. Theory. 45 (2010) 1462–1476. https://doi.org/10.1016/J.MECHMACHTHEORY.2010.05.001.

[15] P. Araujo-Gómez, M. Díaz-Rodríguez, V. Mata, O.A. González-Estrada, Kinematic analysis and dimensional optimization of a 2R2T parallel manipulator, J. Brazilian Soc. Mech. Sci. Eng. 41 (2019) 425. https://doi.org/10.1007/s40430-019-1934-1.

[16] J. Hesselbach, J. Maaß, C. Bier, Singularity prediction for parallel robots for improvement of sensor-integrated assembly, CIRP Ann. - Manuf. Technol. 54 (2005) 349–352. https://doi.org/10.1016/S0007-8506(07)60120-6.

[17] K.. Hunt, E.J.. Primrose, Assembly configurations of some in-parallel-actuated manipulators, Mech. Mach. Theory. 28 (1993) 31–42. https://doi.org/10.1016/0094-114X(93)90044-V.

[18] P. Wenger, D. Chablat, Workspace and Assembly Modes in Fully-Parallel Manipulators: A Descriptive Study, in: Adv. Robot Kinemat. Anal. Control, Springer Netherlands, 1998: pp. 117–126. https://doi.org/10.1007/978-94-015-9064-8_12.

[19] C. Innocenti, V. Parenti-Castelli, Singularity-free evolution from one configuration to another in serial and fully-parallel manipulators, J. Mech. Des. Trans. ASME. 120 (1998) 73–79. https://doi.org/10.1115/1.2826679.

[20] A. Hernandez, O. Altuzarra, V. Petuya, E. MacHo, Defining conditions for nonsingular transitions between assembly modes, IEEE Trans. Robot. 25 (2009) 1438–1447. https://doi.org/10.1109/TRO.2009.2030229.

[21] P.R. McAree, R.W. Daniel, Explanation of never-special assembly changing motions for 3-3 parallel manipulators, Int. J. Rob. Res. 18 (1999) 556–574. https://doi.org/10.1177/02783649922066394.

[22] B. Dasgupta, T.S. Mruthyunjaya, Singularity-free path planning for the Stewart platform manipulator, Mech. Mach. Theory. 33 (1998) 711–725. https://doi.org/10.1016/S0094-114X(97)00095-5.

[23] A.K. Dash, I.M. Chen, S.H. Yeo, G. Yang, Workspace generation and planning singularity-free path for parallel manipulators, Mech. Mach. Theory. 40 (2005) 776–805. https://doi.org/10.1016/j.mechmachtheory.2005.01.001.

[24] S. Caro, P. Wenger, D. Chablat, Non-Singular Assembly Mode Changing Trajectories of a 6-DOF Parallel Robot, in: Vol. 4 36th Mech. Robot. Conf. Parts A B, American Society of Mechanical Engineers, 2012: pp. 1245–1254. https://doi.org/10.1115/DETC2012-70662.

[25] Y. Dai, Y. Fu, B. Li, X. Wang, T. Yu, W. Wang, Clearance effected accuracy and error sensitivity analysis: A new nonlinear equivalent method for spatial parallel robot, J. Mech. Sci. Technol. 31 (2017) 5493–5504.

https://doi.org/10.1007/s12206-017-1044-x.
[26] G. Chen, H. Wang, Z. Lin, A unified approach to the accuracy analysis of planar parallel manipulators both with input uncertainties and joint clearance, Mech. Mach. Theory. 64 (2013) 1–17. https://doi.org/10.1016/j.mechmachtheory.2013.01.005.
[27] N. Binaud, S. Caro, S. Bai, P. Wenger, Comparison of 3-PPR parallel planar manipulators based on their sensitivity to joint clearances, in: IEEE/RSJ 2010 Int. Conf. Intell. Robot. Syst. IROS 2010 - Conf. Proc., 2010: pp. 2778–2783. https://doi.org/10.1109/IROS.2010.5649455.
[28] T. Huang, D.G. Chetwynd, J.P. Mei, X.M. Zhao, Tolerance design of a 2-DOF overconstrained translational parallel robot, IEEE Trans. Robot. 22 (2006) 167–172. https://doi.org/10.1109/TRO.2005.861456.
[29] M. Ohno, Y. Takeda, Identification of Joint Clearances in Parallel Robots by Using Embedded Sensors, IFToMM D-A-CH Konf. Vierte IFToMM D-A-CH Konf. 2018. (2018). https://doi.org/10.17185/DUEPUBLICO/45328.
[30] P. Araujo-Gómez, V. Mata, M. Díaz-Rodríguez, A. Valera, A. Page, Design and kinematic analysis of a novel 3UPS/RPU parallel kinematic mechanism with 2T2R motion for knee diagnosis and rehabilitation tasks, J. Mech. Robot. 9 (2017) 061004. https://doi.org/10.1115/1.4037800.
[31] F. Valero, M. Díaz-Rodríguez, M. Vallés, A. Besa, E. Bernabéu, Á. Valera, Reconfiguration of a parallel kinematic manipulator with 2T2R motions for avoiding singularities through minimizing actuator forces, Mechatronics. 69 (2020) 102382. https://doi.org/10.1016/j.mechatronics.2020.102382.
[32] H. Lipkin, A note on Denavit-Hartenberg notation in robotics, in: Proc. ASME Int. Des. Eng. Tech. Conf. Comput. Inf. Eng. Conf. - DETC2005, 2005: pp. 921–926. https://doi.org/10.1115/detc2005-85460.
[33] X.-J. Liu, J. Wang, Parallel Kinematics, Springer Berlin Heidelberg, Berlin, Heidelberg, 2014. https://doi.org/10.1007/978-3-642-36929-2.
[34] F. Xie, X.J. Liu, Analysis of the kinematic characteristics of a high-speed parallel robot with Schönflies motion: Mobility, kinematics, and singularity, Front. Mech. Eng. 11 (2016) 135–143. https://doi.org/10.1007/s11465-016-0389-7.
[35] G. Chen, W. Yu, C. Chen, H. Wang, Z. Lin, A new approach for the identification of reciprocal screw systems and its application to the kinematics analysis of limited-DOF parallel manipulators, Mech. Mach. Theory. 118 (2017) 194–218. https://doi.org/10.1016/j.mechmachtheory.2017.08.007.
[36] X. Huang, G. He, New and efficient method for the direct kinematic solution of the general planar stewart platform, in: Proc. 2009 IEEE Int. Conf. Autom. Logist. ICAL 2009, 2009: pp. 1979–1983. https://doi.org/10.1109/ICAL.2009.5262630.
[37] M.S. Tsai, T.N. Shiau, Y.J. Tsai, T.H. Chang, Direct kinematic analysis of a 3-PRS parallel mechanism, Mech. Mach. Theory. 38 (2003) 71–83. https://doi.org/10.1016/S0094-114X(02)00069-1.
[38] T. Becker, V. Weispfenning, Gröbner Bases, in: 1993: pp. 187–242. https://doi.org/10.1007/978-1-4612-0913-3_6.
[39] A. Kapandji, Artificial physiology, Editorial médica panamericana, Madrid, 2010.
[40] W.S. Cleveland, C. Loader, Smoothing by Local Regression: Principles and Methods, in: 1996: pp. 10–49. https://doi.org/10.1007/978-3-642-48425-4_2.